\pdfoutput=1
\documentclass{article}

    \PassOptionsToPackage{numbers, compress}{natbib}

\usepackage[preprint]{neurips_2024}

\usepackage[utf8]{inputenc} 
\usepackage[T1]{fontenc}    
\usepackage{hyperref}       
\usepackage{url}            
\usepackage{booktabs}       
\usepackage{amsfonts}       
\usepackage{nicefrac}       
\usepackage{microtype}      
\usepackage{xcolor}         

\usepackage{multirow}
\usepackage{adjustbox}
\usepackage{natbib}
\usepackage{hyperref}

\title{Embodied LLM Agents Learn to \\Cooperate in Organized Teams}

\author{%
  Xudong Guo$^{1}$ \quad  Kaixuan Huang$^{2}$ \quad Jiale Liu$^{3}$ \quad Wenhui Fan$^{1}$ \quad Natalia Vélez$^{2}$ \quad \\\textbf{Qingyun Wu}$^{3}$ \quad \textbf{Huazheng Wang}$^{4}$ \quad \textbf{Thomas L. Griffiths}$^{2}$ \quad \textbf{Mengdi Wang}$^{2}$ \quad\\
  $^{1}$Tsinghua University\quad $^{2}$Princeton University \quad \\
  $^{3}$Penn State University \quad
  $^{4}$Oregon State University\\
}

\begin{document}

\maketitle

\begin{abstract}
\label{sec:abstract}
Large Language Models (LLMs) have emerged as integral tools for reasoning, planning, and decision-making, drawing upon their extensive world knowledge and proficiency in language-related tasks. LLMs thus hold tremendous potential for natural language interaction within multi-agent systems to foster cooperation. However, LLM agents tend to over-report and comply with any instruction, which may result in information redundancy and confusion in multi-agent cooperation. Inspired by human organizations, this paper introduces a framework that imposes prompt-based organization structures on LLM agents to mitigate these problems. Through a series of experiments with embodied LLM agents and human-agent collaboration, our results highlight the impact of designated leadership on team efficiency, shedding light on the leadership qualities displayed by LLM agents and their spontaneous cooperative behaviors. Further, we harness the potential of LLMs to propose enhanced organizational prompts, via a \textit{Criticize-Reflect} process, 
resulting in novel organization structures that reduce communication costs and enhance team efficiency\footnote{Code is available: \href{https://github.com/tobeatraceur/Organized-LLM-Agents}{https://github.com/tobeatraceur/Organized-LLM-Agents}.

Project website: \href{https://organized-llm-agents.netlify.app/}{https://organized-llm-agents.netlify.app/}.}.
\end{abstract}

\section{Introduction}

Modern intelligent systems, such as autonomous vehicle networks and swarms of drones, often involve complex decision-making processes where multiple agents must collaborate seamlessly to achieve specific objectives \cite{wang2020multi, vinyals_grandmaster_2019, zhang_cityflow_2019, wang_hierarchical_2021}. In these systems, communication among the various agents is pivotal, as it dictates the flow of information, coordination of tasks, and overall system performance \cite{zhang2019efficient, ijcai2023p15, foerster2016learning, das2019tarmac}.
Agents in traditional multi-agent systems often have to communicate in pre-specified ways, such as exchanging gradients, sharing data, state observations and actions, etc \cite{kim2020communication, lin2021learning,foerster2016learning}. The emergence of large language models (LLMs) makes it possible for AI agents to communicate and cooperate using natural language, bringing enormous flexibility and potential for more nuanced and human-understandable interactions \cite{park2023generative, hong_metagpt_2023, mandi_roco_2023, chen_agentverse_2023}.

\begin{figure*}[t]
\begin{center}
\centerline{\includegraphics[width=0.9\textwidth]{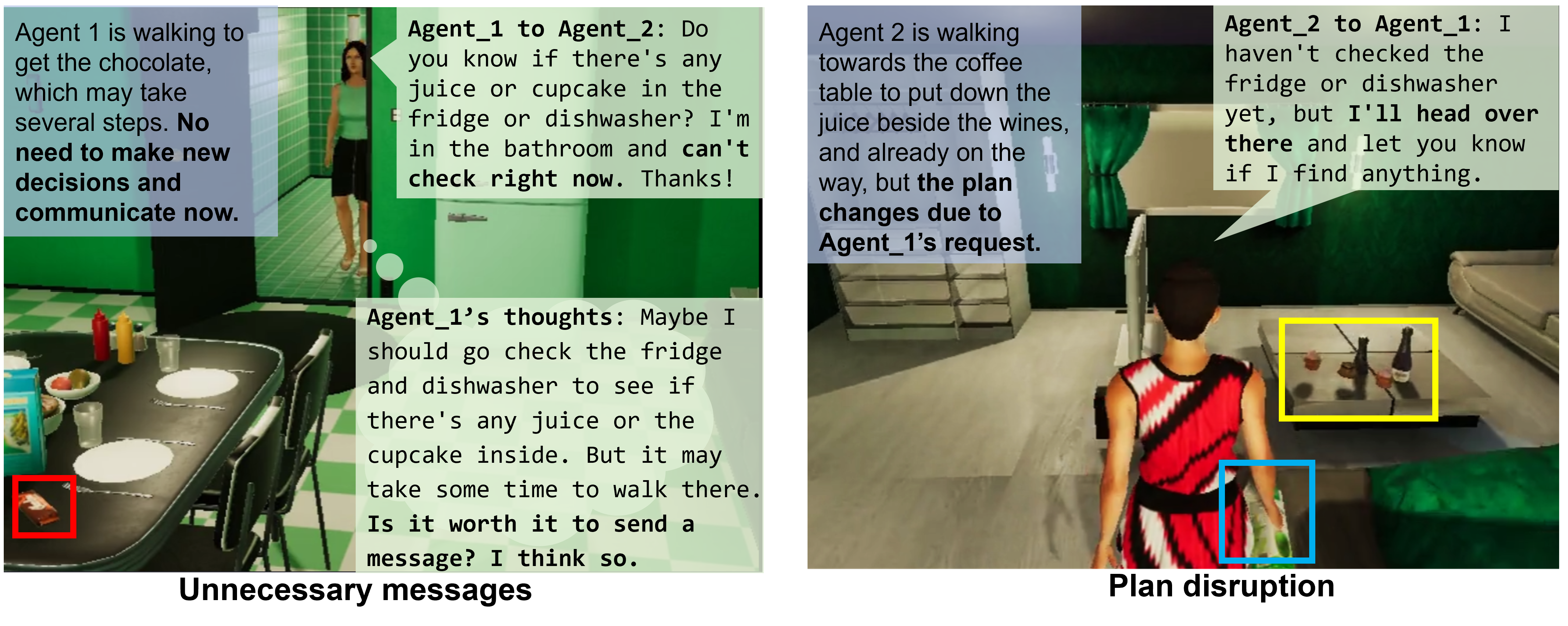}}
\caption{\textbf{Example of disorganized communication and interruption, without a designated leader.} In a team of three GPT-4 agents, two agents engaged in unnecessary communication and made disordered decisions, causing a delay due to the lack of a predefined organization. We identified many more examples including conflicting messages and repetitive communications, see Appendix~\ref{sec:bad:case}. }
\label{bad examples}
\end{center}
\vskip -0.4in
\end{figure*}

Despite the flexibility of LLMs, integrating them into practical multi-agent systems remains a challenge. While LLMs are trained and finetuned for text generation and instruction-following, they are not necessarily tailored to multi-agent cooperation. Modern LLMs are prone to over-reporting and obeying instructions, as a by-product of RLHF finetuning~\citep{bai2022training}, and they can ignore critical information~\citep{liu2023lost} or be distracted by irrelevant information~\citep{shi2023large}, especially when the context is long (see Figure~\ref{bad examples} for examples).
While recent studies involving agent-based LLMs have demonstrated they are capable of solving problems through multi-agent collaboration \cite{li2023theory,zhang_building_2023,mandi_roco_2023}, it is worth noting that such collaborations often follow predefined patterns designed using heuristics to channel the behavior of the models productively \cite{li2023theory}. 
Creating systems that support free-flowing interaction between LLMs in a way that could potentially scale to include humans is still an open problem.

This paper investigates the collaborative potential of LLM agents working in teams. Drawing on prior studies in human collaboration from cognitive and economic perspectives, there is potential for organizations to be redesigned to more effectively manage the limited attention span within teams, as suggested by \citet{simon1971designing}, and mitigate individual limitations and enhance overall team performance, as highlighted by \citet{van1999decentralized} and \citet{velez2023humans}. Specifically, we study two research questions. First, \emph{what role do organizational structures play in multi-LLM-agent systems?} Second, \emph{how can we optimize these organizational structures to support efficient multi-agent coordination?} 
By leveraging AutoGen~\cite{wu2023autogen}, a generic multi-agent conversation framework, we develop a framework for studying how to best organize embodied LLM agents to communicate and collaborate in physical/simulated non-text environments~\cite{zhang_building_2023}. Our framework offers the flexibility to prompt and organize LLM agents into various team structures, facilitating versatile inter-agent communication. It also serves as a testbed to empirically evaluate the traditional ideas proposed in the organization theory literature.

Our initial experiments in this setting reveal that uncoordinated LLM agents often send redundant and repetitive messages and interrupt others' actions, leading to chaos (see Fig.~\ref{bad examples} and Appendix~\ref{sec:bad:case}). To remedy these issues, we explore organizational structures, i.e., the dynamics of information exchange, that allow multiple LLM agents to collaborate and complete a common task efficiently. 

The first organizational structure we explore is a {\em hierarchy}, a classic object of study in organizational theory  \cite{march1958organizations,radner1993organization,chisholm1992coordination,bolton1994firm,garicano2000hierarchies,dodds2003information}. With a designated leader, LLM agents work more efficiently and collaboratively. For the example of a three-agent team, imposing a leader improves efficiency by up to 30\% with almost no extra communication cost (up to 3\%), consistent with findings for human organizations \cite{dodds2003information}. This also holds true in five-agent cases. Further, LLM agents demonstrated the potential to elect their own leader and adjust leadership dynamically via communication. 
With proper organizations, LLM agents exhibit a variety of cooperative behaviors that mimic humans. For example, agents can provide constructive suggestions and seek help from others; they can also execute appropriate interactions for a hierarchy such as reporting back on task progress; see Figures \ref{cooperation examples}, \ref{Behavior ratio} and Appendix~\ref{sec:good:case}. 
We also tested human-agent collaboration, and observe that, unsurprisingly, human leaders are much better at coordinating a team of agents when compared to AI agents.

In addition to testing existing organizational structures, we explore the use of LLMs to improve the organizational prompts. To this end, we develop a {\em Criticize-Reflect} framework, adopting a dual LLM architecture,  to reflect on the team performance and generate improved and novel organizational prompts. Through this iterative process, our LLM agents spontaneously form novel, effective team structures, leading to reduced communication cost and improved efficiency; see Figures \ref{Iteration of optimization} and \ref{comm fig}.

To summarize, our main contributions are: 1. We design a novel multi-LLM-agent architecture for $\geq 3$ embodied agents, facilitating flexible communication to implement emergent organizational structures. 2. We develop a \textit{Criticize-Reflect} framework based on LLMs to improve the organizational prompts automatically. 3. Extensive experiments demonstrate that hierarchical organization improves team efficiency, which aligns well with existing literature on human organizations.

\section{Related Works}

\subsection{LLM Agents}

As powerful LLMs inherit abundant world knowledge and also general reasoning ability, there are increasing efforts to deploy LLMs as the reasoning core for decision-making to build human-like autonomous agents~\citep{sun2023adaplanner, zhu2023ghost, hao2023reasoning}.
This requires observations of the RL environment to be translated into natural language in a way that is easier for LLMs to process. 
The reasoning of the LLMs also needs to be turned into a viable action for execution. Popular prompting techniques for doing so include ReAct~\citep{yao2022react} and Reflexion~\citep{shinn2023reflexion}. Other methods that involve fine-tuning the language models have also been explored~\citep{hao2023toolkengpt}.
In addition, various techniques have been proposed to mitigate the biases and constraints of LLMs, including chain-of-thought reasoning~\citep{wei2022chain}, external tools~\citep{shen2023hugginggpt, patil2023gorilla}, external documents~\citep{wang2023voyager} and skill libraries~\citep{zhu2023ghost}.

\subsection{Multi-Agent Cooperation}

Multi-agent cooperation has been extensively studied for decades under various topics such as communication efficiency, planning, leadership, and team dynamics using different platforms~\citep{lowe2017multi,samvelyan2019starcraft,resnick2018pommerman,puig_watch-and-help_2021} (see recent surveys for detail~\citep{oroojlooy2023review,zhang2021multi,gronauer2022multi}). Previous works mainly focused on communication through continuous vectors~\citep{das2019tarmac} or discrete symbols~\citep{lowe2017multi,jaques2019social}. Recent works~\citep{xu2023language, zhang2023exploring, wu2023autogen,li2023camel, ishibashi2024self, li2023metaagents, talebirad2023multi} showed that multiple LLM agents or human-agent
teams can improve upon single LLM in solving pure text-based tasks, such as creative writing, reasoning, and code generation. Other works~\citep{liu_dynamic_2023, hong_metagpt_2023, zheng2023agents} further explored agent selection or role assignment to improve the performance.

LLMs have also been applied to multi-agent cooperation for embodied tasks~\citep{agashe_evaluating_2023, mandi_roco_2023, park2023generative, chen_agentverse_2023}. Besides, \citet{zhang_proagent_2023} proposed an intention inference framework to enhance the cooperation of LLM agents without explicit
communication. \citet{li2023theory} investigated LLM-agents collaboration for Theory of Mind inferences tasks with a broadcast-only communication protocol and homogeneous policies. \citet{zhang_building_2023} studied embodied multi-agent cooperation in the two-agent and the one-human-one-agent settings. \citet{chen_scalable_2023} explored different fixed communication structures for multi-LLM-robots. \citet{zhao2024hierarchical} and \citet{chen2024s} organized the agents by predefined and fixed communication with a virtual manager. These initial explorations are limited to fixed team structures and are not optimized for communication efficiency. 
In contrast, our work explores the impact of deploying and optimizing organizational structures, allowing $\geq 3$ agents in a team, for efficient multi-agent communication and cooperation.

\subsection{Prompt Optimization}

Language models are sensitive to prompts. The format of the prompt can have a substantial influence on performance~\citep{gao2020making, wei2022chain, zhou2022least, shi2023large, zou2023universal, qi2023visual}. Various research efforts have aimed at prompt optimization. Typical approaches include heuristic search using language models' knowledge~\citep{gao2020making, shin2020autoprompt}, first-order methods like soft prompt tuning~\citep{lester2021power}, and prefix tuning~\citep{li-liang-2021-prefix}. In this work, we focus on obtaining an interpretable prompt in the form of natural language, drawing on insights from \citet{yang2023large}, \citet{zhou2022large}, and \citet{pryzant2023automatic}.

\section{Method}
\subsection{Architecture and Multi-Agent Communication}

\begin{figure*}[t]
\begin{center}
\centerline{\includegraphics[width=0.9\textwidth]{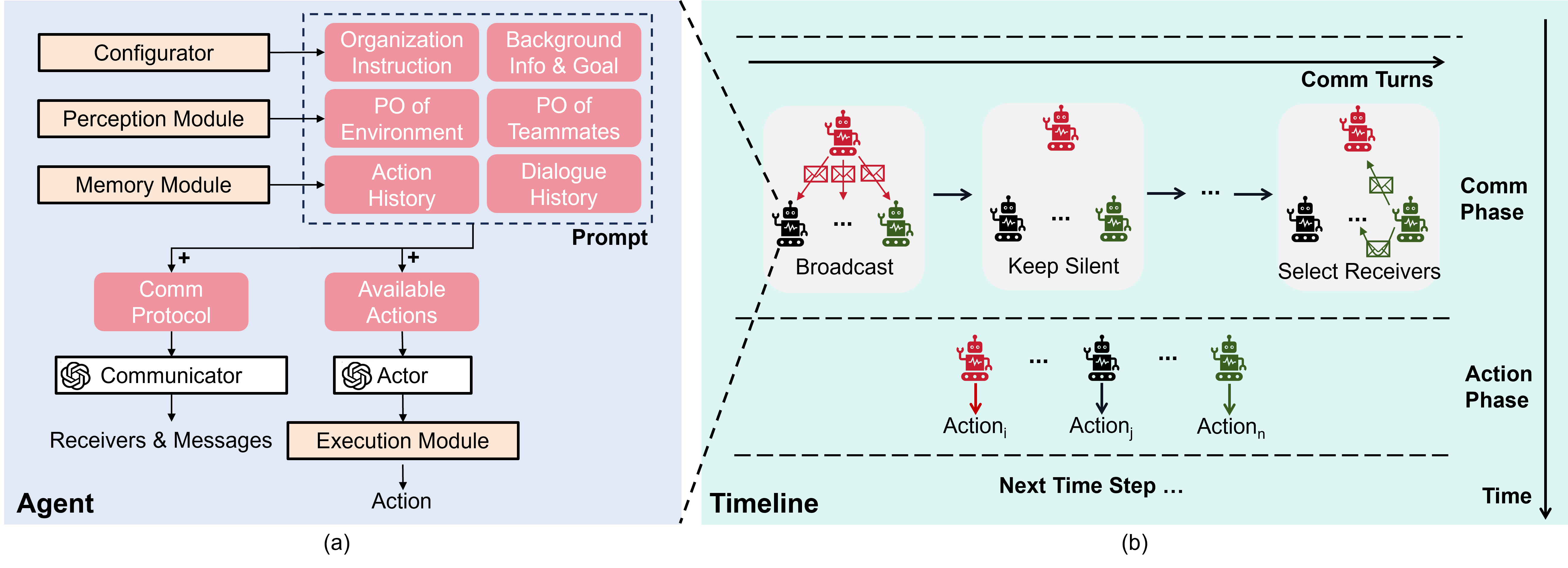}}
\caption{\textbf{Multi-LLM-agent architecture.} (a) The modules of an LLM agent and the composition
of prompts. (b) There are two phases in one time step: Communication phase and Action phase. In the communication phase, the agents take turns communicating by broadcasting or selecting receivers to send distinct messages. The agents can also choose to keep silent. Comm is short for Communication; PO is short for Partial Observation.}
\label{Architecture}
\end{center}
\vskip -0.2in
\end{figure*}

We adopt the embodied LLM-agent architecture proposed by \citet{zhang_building_2023} and expand it to enable organized teams of $\geq 3$ agents to communicate, plan, and act in physical/simulated environments. Figure~\ref{Architecture} illustrates our architecture. Borrowing insights from \citet{zhang_building_2023}, we adopt four standard modules: Configurator, Perception Module, Memory Module, and Execution Module. They are responsible for configuring the agents, translating environmental observations into text, storing \& retrieving historical information, and executing actions, respectively (Fig.~\ref{Architecture}(a)). 

Previous works focused on two-agent cooperation, in which case the communication can be simply treated as an extra action \cite{mandi_roco_2023, zhang_building_2023}. In contrast, we aim to enable three or more agents to work in a team and cooperate through emergent organized communication. Thus we design the architecture with several features that facilitate organized multi-agent communication  (Figure~\ref{Architecture}(b)): 
\begin{itemize}
\item We disentangle the communication decision-making from the action decision-making by adopting two separate LLMs as Actor and Communicator.
\item We impose an organizational structure for the agent team via prompting, i.e., including a textual description as part of the prompts for both the Actor and Communicator.
\item
LLM agents keep alternating between two phases during their task: the communication phase and the action phase.  The standalone communication phase supports richer team structures and flexible communication patterns.
\item During communication, agents take turns to communicate. An agent can choose to broadcast a message, select one recipient for a message, choose multiple recipients and send them distinct messages, or remain silent. 
Agents keep their own history of communication and can respond to messages from previous communications. 
\end{itemize}

\subsection{\textit{Criticize-Reflect} Method for Improving Organizational Structure}
\label{LLM-designed instrctions}

\begin{figure*}[t]
\begin{center}
\centerline{\includegraphics[width=0.85\textwidth]{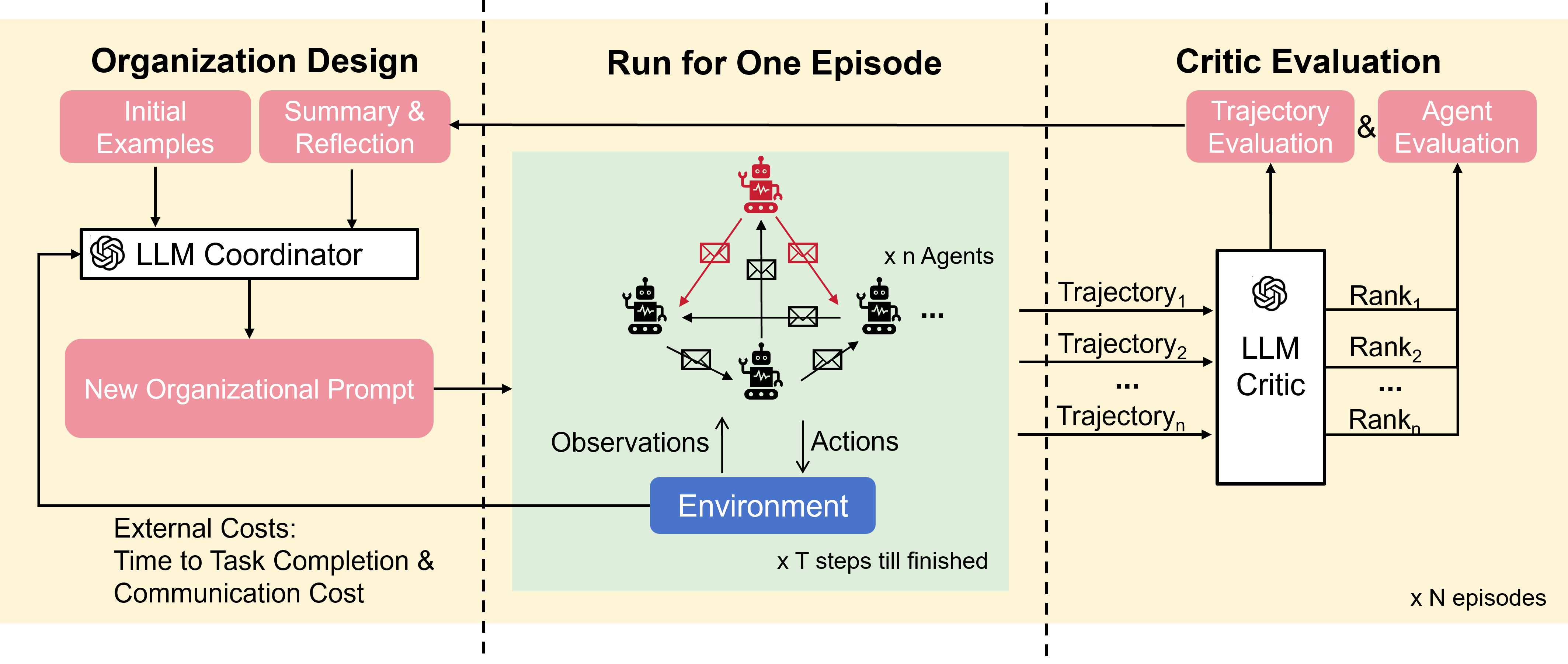}}
\caption{\textbf{\textit{Criticize-Reflect} architecture for improving organizational structure.} The red agent represents the leader in a hierarchically-organized team. After the team completes one episode, the Critic evaluates the trajectories and analyzes the agents' performance. Together with the external costs from the environment, the Coordinator proposes a new organizational prompt to improve the team efficiency. The new prompt will be applied to the next episode to continue the iteration. 
}
\label{Optimization framework}
\end{center}
\vskip -0.2in
\end{figure*}

We leverage powerful LLMs to optimize the organizational prompt, borrowing insights from~\citep{yang2023large}. To do so, we introduce a dual-LLM framework to allow the multi-LLM-agent system to ponder and improve the organizational structure. 
Figure~\ref{Optimization framework} illustrates the architecture of our framework. It consists of two LLMs:
\begin{itemize}
    \item {\bf LLM critic:} Inspired by the Actor-Critic method of reinforcement learning \cite{konda1999actor}, we introduce an LLM critic to evaluate the team's performance based on verbal feedback. The team critic takes as input the dialogue and action history of one episode. Then, the critic analyzes the input and reasons to extract and summarize the key steps that are believed to influence the performance. Also, the critic provides a textual evaluation of agents' behaviors and the ranking of their leadership. See the prompts in Appendix~\ref{sec:prompt} and technical details (including the ranking criteria) in Appendix~\ref{sec:critic:outputs}.
    \item {\bf LLM coordinator:} The LLM coordinator takes as input the outputs of the LLM critic as well as cost metrics (time to task completion and communication cost) of previous episodes from the environment. It reflects on these data and generates thoughts based on the analysis of the past episodes  and the initial examples. With the reflection of organizational prompts and their performance, the coordinator proposes a new and different organizational prompt for the next episode.  
    Please refer to Appendix~\ref{sec:prompt} for the prompts and Appendix~\ref{sec:definitions} for the details of reflection.
\end{itemize}
    For each new organizational prompt, we run for one episode and then return the dialogue and action history to the critic. By criticizing and reflecting on the prompts iteratively, the framework discovers more effective, novel organizational structures with \emph{self-improvement}.

\subsection{Environment Setup}
\label{sec:setup}
We chose VirtualHome-Social \cite{puig2018virtualhome, puig_watch-and-help_2021} as the environment and extended it to support multi-LLM-agent communication and interaction. In this environment, agents are humanoid helpers in a virtual home doing housekeeping, where the tasks include \textit{Prepare afternoon tea, Wash dishes, Prepare a meal, Put groceries, Set up a dinner table}, etc. For instance, in Figure~\ref{bad examples}, the agents cooperate to prepare afternoon tea by searching for and transporting task-specific items (chocolate, juice, wine, etc.) to a target location (the coffee table).
The environment generates symbolic observations of the objects in the home and their relations. Each agent only observes the objects in the open containers located in her room and teammates in the same room, but she can walk to another room to explore. Any agent can communicate with any other agent, not subject to a range limit.

Each episode starts from an initial state where agents are randomly located in the environment and all containers are closed. The episode terminates when the task is fully completed. 
To evaluate the team's efficiency we measure the number of time steps taken to task completion, and we report the average number of tokens communicated between agents per step. In our experiment, each run initializes with an independently randomized state to obtain the mean and a confidence interval. 
We adopt GPT-4, GPT-3.5-turbo~\citep{ouyang2022training}, and Llama2-70B~\citep{touvron2023llama} as LLMs in our agents. 
The temperature is set as 0.8, the maximum number of output tokens is 256, and the number of completion choices to generate is 1. In practice, we use two Nvidia 80GB A100 GPUs to do the inference on Llama2-70B.

\section{Main Results}
\label{sec:results}
\begin{figure*}[t]
\begin{center}
\centerline{\includegraphics[width=1.075\textwidth]{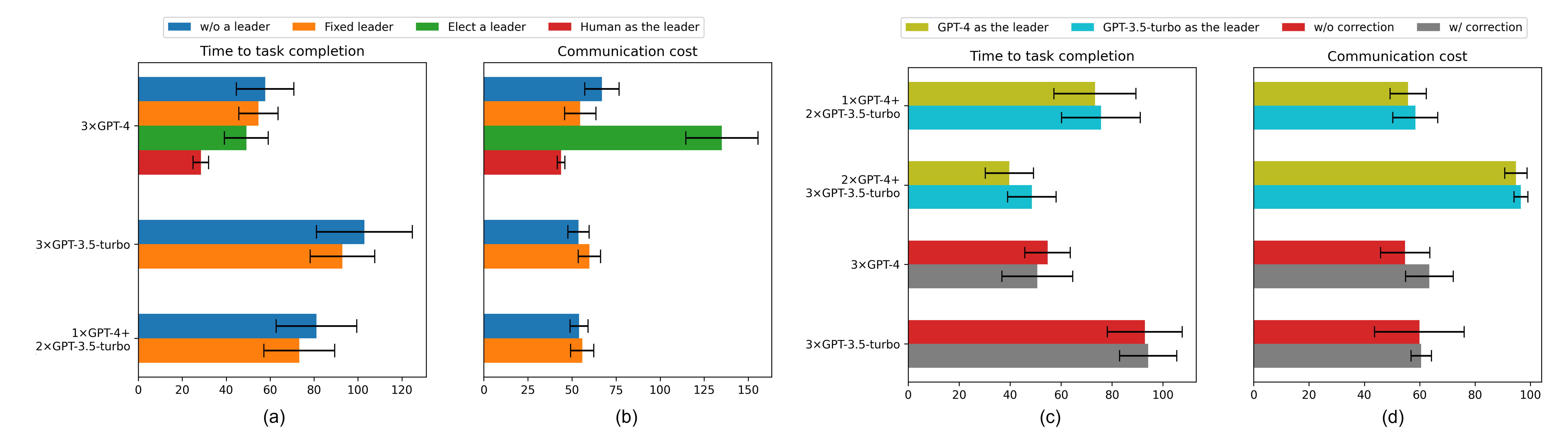}}
\caption{\textbf{Organized teams with a designated leader achieve higher efficiency.} (a,b) Comparison between the case of disorganized agents, the case where a leader is appointed, the case where agents choose their own leader dynamically, and the case where a human player replaces an agent to be the leader. Note that GPT-3.5-turbo doesn't support leadership election. 
(c,d) Comparing leadership quality for GPT-3.5-turbo vs. GPT-4. The confidence intervals of Human as the leader group are calculated over 3 seeds while others are over 20 seeds.}

\label{results fig}
\end{center}
\vskip -0.2in
\end{figure*}

\begin{figure*}[t]
\begin{center}
\centerline{\includegraphics[width=0.8\textwidth]{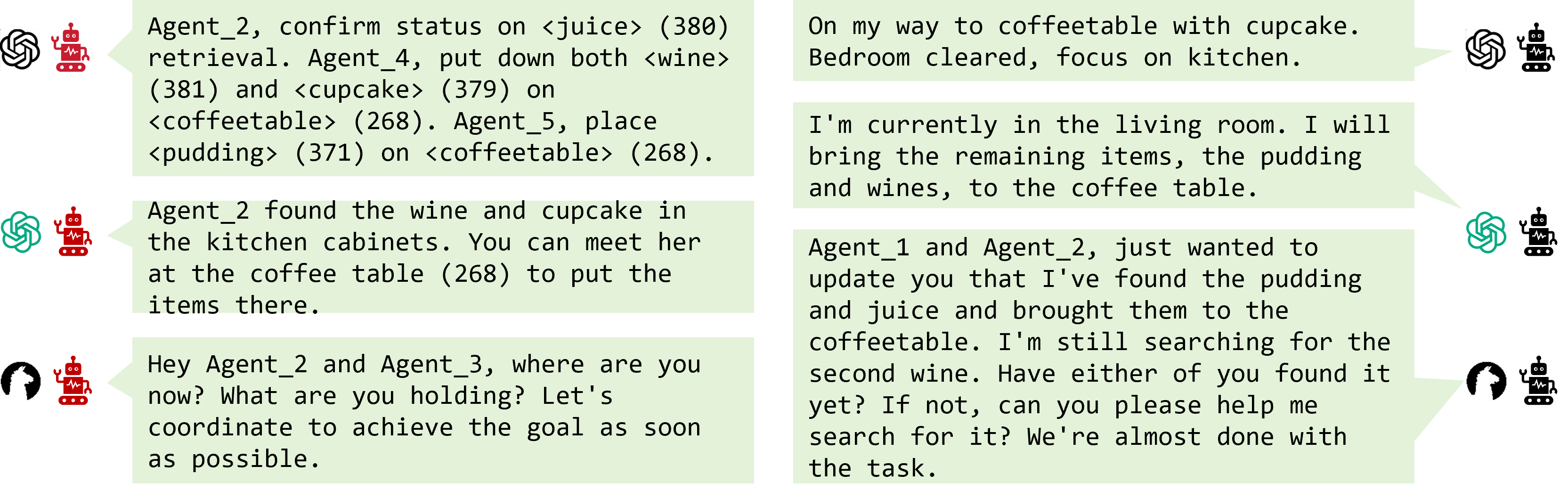}}
\caption{\textbf{Examples of communication messages when there is a designated leader.} Left: messages from lead agents; Right: messages from non-lead agents. GPT-4 (upper), GPT-3.5-turbo (center), and Llama2-70B (lower) demonstrated different communication styles.}
\label{message examples}
\end{center}
\vskip -0.4in
\end{figure*}

\subsection{A Designated Leader Enhances  Performance}

We first studied the effect of organizational structures and leadership on LLM agents. For benchmarking, we experimented with disorganized LLM agents without providing any organizational prompt. In this case, agents still communicate with one another and work to complete the overall task. However, we discovered frequent occasions where agents send redundant, repetitive messages and interfere with one another. See Figure~\ref{bad examples} for an illustration and see Appendix~\ref{sec:bad:case} for more examples. Numeric metrics are reported in Appendix Table~\ref{results table}. 

When a leader is appointed via the organizational prompt, we observe improved team performance -- the teams completed the task in less time (Figure  \ref{results fig}(a)). After running the 3$\times$GPT-3.5-turbo experiments with 20 random seeds, we performed two-sample t-tests, showing a statistically significant improvement by 9.76\% in performance ($t(38) = 1.71, p < .05$). Similarly, a designated leader brings benefits to the team of 3$\times$GPT-4 (improved by 5.28\%, $t(38) = 0.86, p = 0.20$) and the team of 1$\times$GPT-4+2$\times$GPT-3.5-turbo (improved by 9.61\%, $t(38) = 1.43, p = 0.08$). Compared to the disorganized teams,  teams with a designated leader only have a slightly increased or even less communication cost (Figure \ref{results fig}(b)). This is consistent with patterns seen in previous models of hierarchical organizations \cite{dodds2003information}. Teams with a leader also emerge centralized communication patterns shown in Figure~\ref{comm fig} and Appendix~\ref{sec:pyramid}. For additional experiments on Llama2-70B, please see Appendix Table~\ref{results table}.
The communication styles of leaders and non-leaders were clearly differentiated, as shown in Figure \ref{message examples}. We further scaled up the team sizes and found that the communication costs only increased in a nearly linear way, without a curse of dimension (See Appendix Table~\ref{tab:scale}).

 Next, we asked the agents to elect their own leader. The leadership was reelected about every 9 time steps, based on information extracted from the latest 12 messages.  
 We observe that agents are generally not power-seeking: they often vote for others to lead. In some occasions,
 agents favored candidates who exhibited higher knowledge levels, for example, one agent thought that "\textit{Given that Agent\_2 has found a necessary item, it makes sense for him to be the leader in this round.}" However, on most occasions, we could not tell whether agents made their votes based on rational reasoning or just random thoughts (see Appendix~\ref{sec:election}). In the case of the 3$\times$GPT-4 team\footnote{Note that GPT-3.5-turbo agents do not support election probably due to their alignment policy and always ignore the demand of election.}, implementing leadership election resulted in improved team efficiency when compared to consistently following a predetermined leader ($t(38) = 1.84, p < .05$; see Figure \ref{results fig}(a)). However, this improvement was accompanied by a substantial increase in communication cost, akin to real-world scenarios where relaxing hierarchical structure potentially increases communication cost \cite{malone2004future}.

The proposed multi-LLM-agent architecture is also human-friendly to support \emph{human-AI collaboration}. In the experiment, we ask a human player to replace the leader in the team of 3 GPT-4 agents. We recruit three human players to conduct the experiments. Figure~\ref{results fig}(a, b) demonstrates that human leadership achieved better task completion time and improved communication efficiency compared with GPT-4 as the leader. Please find more examples of dialogues between the human leader and LLM agents in Appendix~\ref{sec:human-AI}.

\subsection{Leadership and Open Communication Matters}

 LLM agents have different levels of leadership. In the team with a mixture of GPT-4 and GPT-3.5-turbo agents, appointing GPT-4 as the leader increases the team efficiency higher than if GPT-3.5-turbo is the leader (Figure~\ref{results fig}(c,d), Appendix~\ref{sec:leadership}). We ran this experiment on teams of three agents and five agents, respectively. In both scenarios, the task completion time and communication cost are reduced when GPT-4 acts as the leader. This finding implies different levels of leadership between these LLMs.

We also observed that encouraging constructive feedback to the leader agent helped performance. Motivated by successful human organizations, we tried to promote open communications among LLM agents by adding an additional prompt that \textit{"If the leader's instructions are not right, you can correct the leader"}. Figure \ref{results fig}(c, d) illustrates the results. Interestingly, this modification improves the team's overall efficiency and reduces the time to task completion when the team is made up of 3$\times$GPT-4 ($t(38) = 0.87, p = 0.14$). In contrast, the same modification lowers the team efficiency when GPT-3.5-turbo agents try to correct the leader ($t(38) = 0.27, p = 0.40$). In both experiments, the communication cost increases. We present more details about these behaviors in Appendix~\ref{sec:correction}.

\subsection{Emergence of Cooperative Behaviors}

\begin{figure*}[t]
\begin{center}
\centerline{
\includegraphics[width=1.05\textwidth]{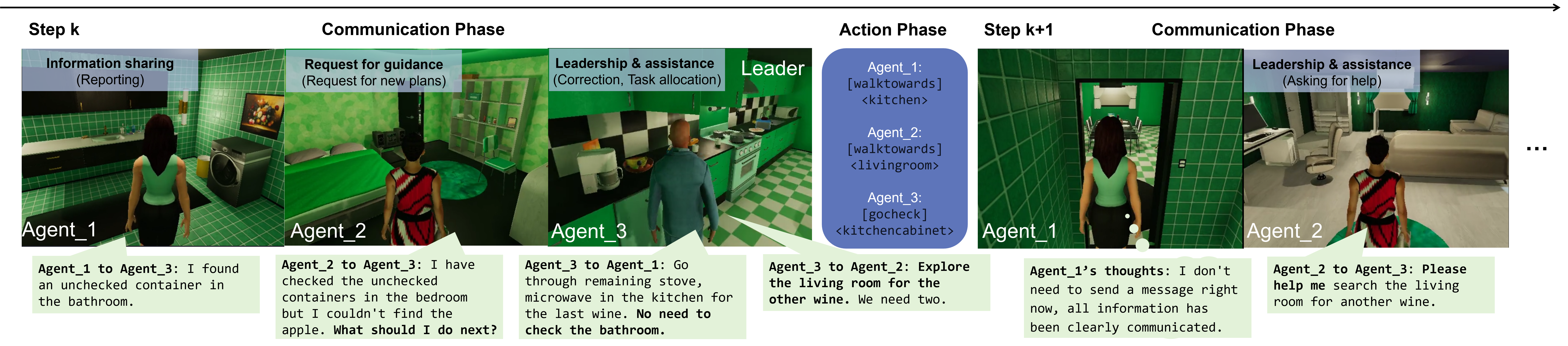}}
\caption{\textbf{Examples of cooperative behaviors in a dialogue.} Agent\_3 leads the team (3$\times$GPT-4 agents). The agents emerge three types of cooperative behaviors: information sharing, leadership \& assistance, and request for guidance.
}
\label{cooperation examples}

\end{center}
\vskip -0.2in
\end{figure*}

\begin{figure*}[t]
\begin{center}
\centerline{
\includegraphics[width=0.8\textwidth]{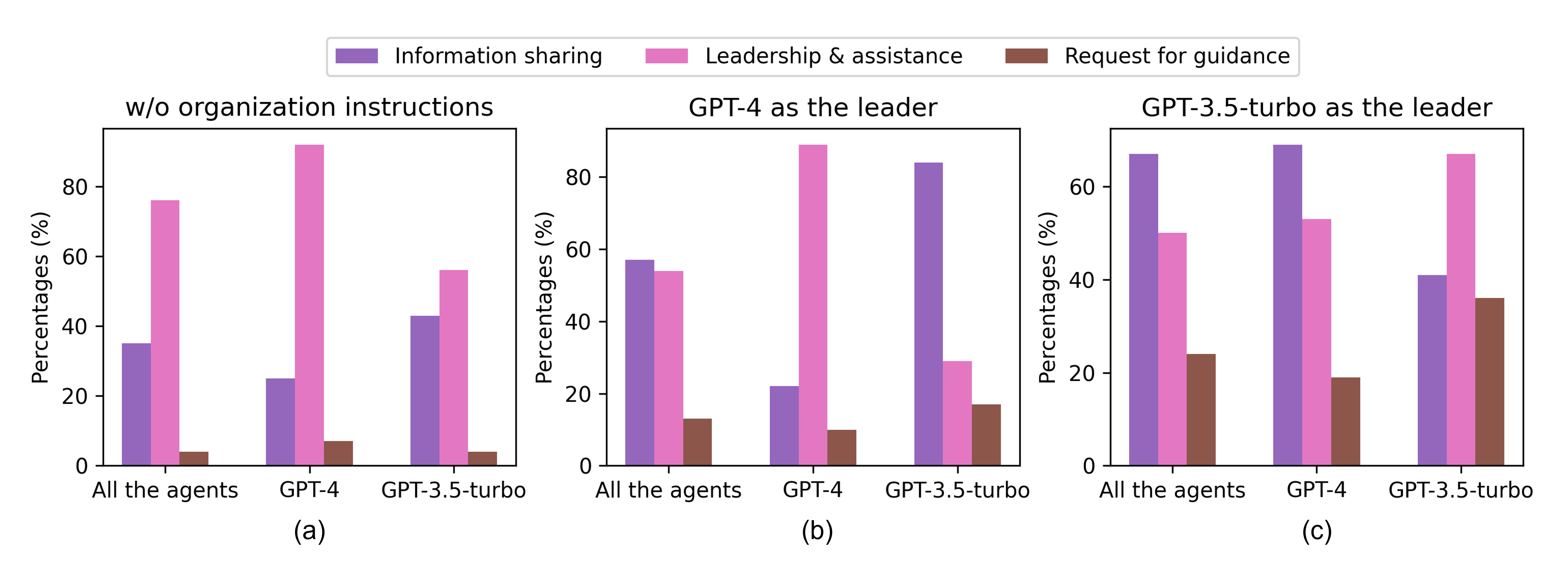}}
\caption{
\textbf{Emergent cooperative behaviors of LLM agents.} We analyzed the communication log of the mixture team (1$\times$GPT-4+2$\times$GPT-3.5-turbo) and asked another GPT-4 to annotate agent's cooperative behaviors. (a) Behavior of disorganized agents. (b) Behavior of a team led by a GPT-4 agent. (c) Behavior of a team led by a GPT-3.5-turbo agent.
}\label{Behavior ratio}
\end{center}
\vskip -0.2in
\end{figure*}

We delved into the behaviors of LLM agents in an organized team to investigate how organizational prompts influence agents' communication and decisions. Analysis of their dialogue history revealed that agents demonstrated a variety of cooperative behaviors, such as reporting, correction, task allocation, and asking for help (see Figure \ref{cooperation examples} for an example dialogue). 

One may argue that these types of behaviors could also emerge due to the nature of LLMs, even without a pre-specified team structure. 
Thus we performed a quantitative analysis to study the impact of an organizational prompt on these behaviors. We followed a three-step process:
\begin{itemize}
    \item[(1)]  We defined three major categories of human cooperative behaviors: (i) Information sharing: agents influence others by offering new information, either actively or by being asked. Reporting to the leader, sharing new observations, and answering questions belong to this category. (ii) Leadership \& assistance: agents, especially the leader if there is one, can influence others by changing their plans. The behaviors include task allocation, correction, and asking for help. (iii) Request for guidance: agents actively request new information or plans for their own decision-making.

    \item[(2)]  We developed a standalone prompt-based GPT-4-classifier to analyze each piece of dialogue. The classifier decides whether to label the dialogue with any subset of the aforementioned labels. The classifier has an accuracy of 91.67\% when tested on 20 human-labeled dialogue samples with 60 labels (see Appendix~\ref{sec:prompt} for the prompt and Appendix~\ref{sec:classification} for the test samples).

    \item[(3)] We use the classifier to label messages generated by the agents and report the percentages of messages with cooperative behaviors. Note that one message may have multiple labels.
\end{itemize}

Figure \ref{Behavior ratio} reports the results and illustrates the behavior patterns for different LLM agents. 
The results support several observations. Even in a disorganized team, LLM agents love to tell others what to do. Leadership \& assistance accounts for around $>$ 50\% of all the behavior (Figure \ref{Behavior ratio}(a)). However, other than telling others what to do, agents in the disorganized team do not show much cooperative behavior, for example, they would request for guidance in $<10\%$ of the dialogues. 

In contrast, when the team has a hierarchical organization,
the lead LLM agent would presume a dominant role and give orders to others (amount to $>60\%$ of their communication), while other members tend to follow and give fewer orders compared with the disorganized case.  (Figure~\ref{Behavior ratio}(b, c)). In such a team, agents tend to share and ask for more information, especially for the follower agents in the team. But still, the agents may fail to cooperate well, such as being lazy and confused about numbers, please see examples in Appendix~\ref{sec:failure}.

\subsection{Novel Organizational Structures} 
\label{sec:novel:structure:main}

Having evaluated the merits of different kinds of structures, we let the LLMs propose novel organizational structures and iteratively refine the organizational prompts using the \textit{Criticize-Reflect} method discussed in Section 3 (see also Figure \ref{Optimization framework}).

\begin{figure*}[t]
\begin{center}
\centerline
{ \includegraphics[width=\textwidth]{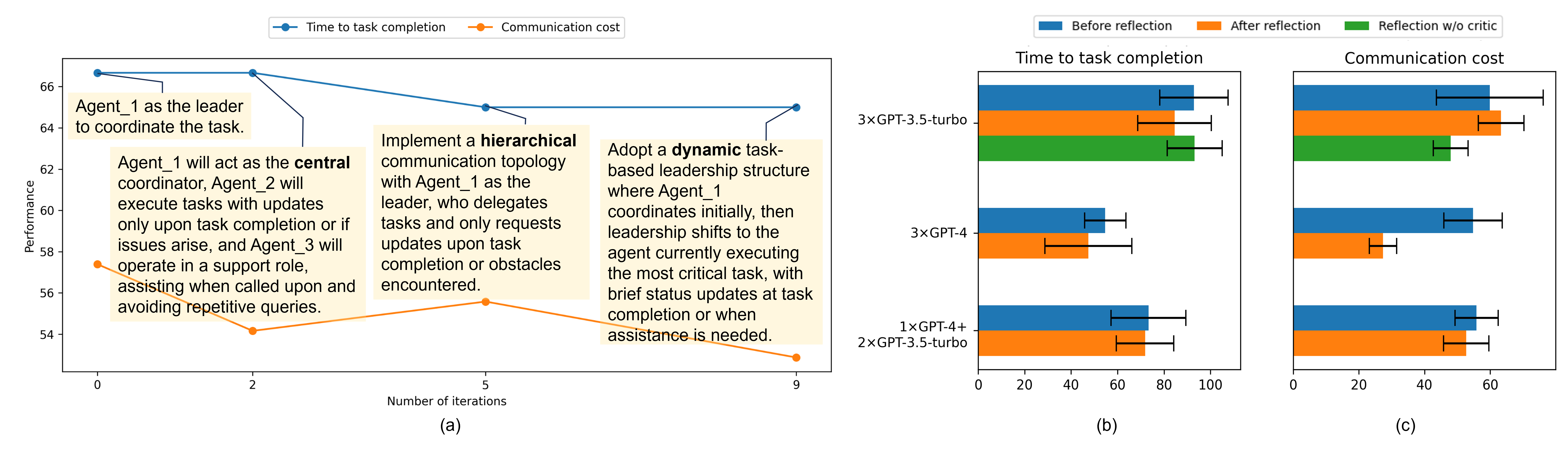}}
\caption{\textbf{The reflection and improvement process for finding novel organizational structures}. 
 (a) The experiment was done using the  1$\times$GPT-4+2$\times$GPT-3.5-turbo team. The organizational prompt evolves during the iterations, and takes on additional keywords such as \textit{"central", "hierarchical"}, and \textit{"dynamic"}.
(b) The confidence intervals are calculated over 20 seeds. 
}
\label{Iteration of optimization}
\end{center}
\vskip -0.4in
\end{figure*}

Figure \ref{Iteration of optimization}(a) visualizes the reflection process. 
The system was initialized with a basic organizational prompt, i.e., {\centering \textit{"Agent\_1 as the leader to coordinate the task"}. }As the Reflection process moves forward, the Coordinator generates a sequence of evolving organizational prompts, picking up key words like "hierarchical" and "dynamic" that imply more complex team structures. 
 
We compared the team's performance before and after the \textit{Criticize-Reflect} steps. Figure \ref{Iteration of optimization}(b) illustrates the team's efficiency. We observe that for 3$\times$GPT-3.5-turbo, the new organizational structure improved the team's efficiency in completing the task ($t(38) = 1.73, p < .05$), at slightly increased communication cost. While for 3$\times$GPT-4 and 1$\times$4+2$\times$GPT-3.5-turbo, the communication cost is reduced with improved task efficiency ($t(38) = 1.56, p = 0.06$ for 3$\times$GPT-4, and $t(38) = 0.32, p = 0.38$ for 1$\times$4+2$\times$GPT-3.5-turbo).

The Critic analyzes the records of action and dialogue, and performance metrics from the most recent episode. It provides evaluation for the full team's trajectory, feedback to individual agents and their rankings. See the example of the Critic outputs in Appendix \ref{sec:critic:outputs}.

As an ablation study, we removed the Critic from our architecture and only performed the Reflection step. The results are shown in Figure \ref{Iteration of optimization}(b), indicating that Reflection without the Critic leads to performance decline ($t(38) = 1.96, p < .05$). In this case, the Coordinator needs to digest all dialogue history and generate a new organizational prompt. This did not work  well and led to rather vague outcomes, for example, \textit{"Establish a flexible communication network with rotating leadership roles assigned based on agents' task-specific expertise to facilitate swift decision-making and reduce unnecessary communication steps."} This comparison highlights the role of the Critic and the importance of having a dual \textit{Criticize-Reflect} architecture. For more results/prompts generated by the reflection process, please refer to Appendix~\ref{sec:refelction}.

\begin{figure*}[t]
\begin{center}
\centerline{\includegraphics[width=\textwidth]{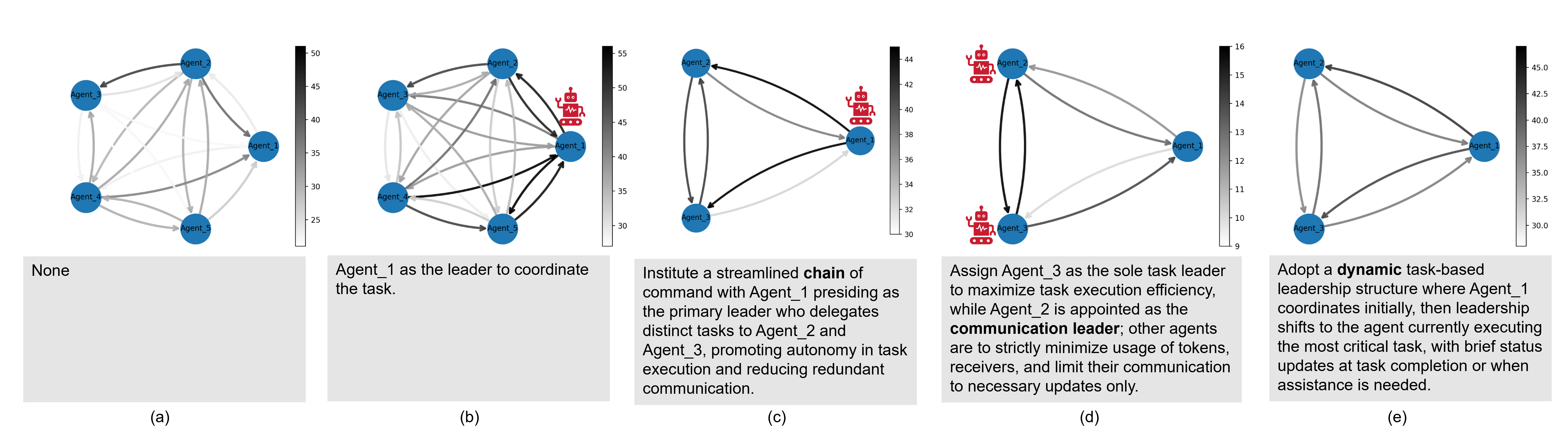}}
\caption{\textbf{Communication patterns and the corresponding organizational prompts.} (a) Team without organizational prompts. (b) Team with a leader. (c) A team in the chain structure. (d) A dual-leader team. (e) A team with a dynamic leadership. (c, d, e) are proposed by \textit{Criticize-Reflect}. Red-robot nodes mark the lead agents, and other nodes are the followers. Edges mark the accumulated communication cost between the two nodes (darker edge means higher token cost).} 
\label{comm fig}
\end{center}
\vskip -0.3in
\end{figure*}

In addition, it is worth mentioning that LLMs are able to generate highly complex prompts that imply novel organizational structures that are rarely seen in human societies. We illustrate the communication patterns as team structures in Figure~\ref{comm fig} together with the three novel structures proposed by \textit{Criticize-Reflect}: (c) chain, (d) dual-leader, and (e) dynamic structures, which are the best structures of the three settings in Figure~\ref{Iteration of optimization}(b, c) respectively.

Finally, to test the generalizability of the novel organizational structures, we pick the best novel prompt, the one illustrated in Figure~\ref{comm fig}(e), proposed by the \textit{Criticize-Reflect} architecture on the \textit{Prepare afternoon tea} task. We test it on a set of six new tasks, comprising of three easy tasks and three hard tasks\footnote{The hard tasks have typical numbers of steps to accomplish the tasks $>60$, while those of easy tasks are $<60$.}, as shown in Appendix Figure~\ref{Across_tasks}. In the three hard tasks, the team with the novel organizational structure had better performances than the team appointing a fixed GPT-4 agent as the leader (Appendix Figure~\ref{Across_tasks}(a,b)). In the three easy tasks, the benefits are marginal. We compared the two teams across all tasks, performed a t-test and concluded that the novel team structure leads to more efficient performance than the fixed leader ($t(22) = 2.08, p < .05$).

\section{Conclusion}
\label{sec:conclusion}
We develop a novel multi-LLM-agent architecture to facilitate communication and organize the embodied agent teams for enhanced cooperation. Moreover, we propose the \textit{Criticize-Reflect} framework based on LLMs to generate more efficient organizational
prompts. Extensive experiments with various group settings and organizational structures demonstrate that a hierarchically-organized team with a designated/elected leader has superior team efficiency, which can be further improved by \textit{Criticize-Reflect}.

 The current work is performed in a single environment and lacks human evaluation. Future work shall extend to a broader set of environments, allowing human evaluation. As VirtualHome cannot hold hundreds of agents, future work can also explore larger organizations in other environments.


\bibliography{main}
\bibliographystyle{plainnat}

\newpage
\appendix
\onecolumn

\section{Prompt Templates}
\label{sec:prompt}
We list the prompts of Actor, Communicator, Critic, and Coordinator as follows.

\textbf{Actor and the Communicator.} \texttt{ORGANIZATION\_INSTRUCTION} is the placeholder for the organization instruction prompt, either manually designed or automatically generated. The environment will provide text descriptions for the current \texttt{GOAL}, \texttt{PROGRESS}, and \texttt{AVAILABLE\_ACTIONS}.
We include the latest $12$ sent and received messages as \texttt{DIALOGUE\_HISTORY}, and the latest $20$ steps of actions as \texttt{ACTION\_HISTORY}.

\vskip 0.2in
\begin{center}
\centerline{
\includegraphics[width=\columnwidth]{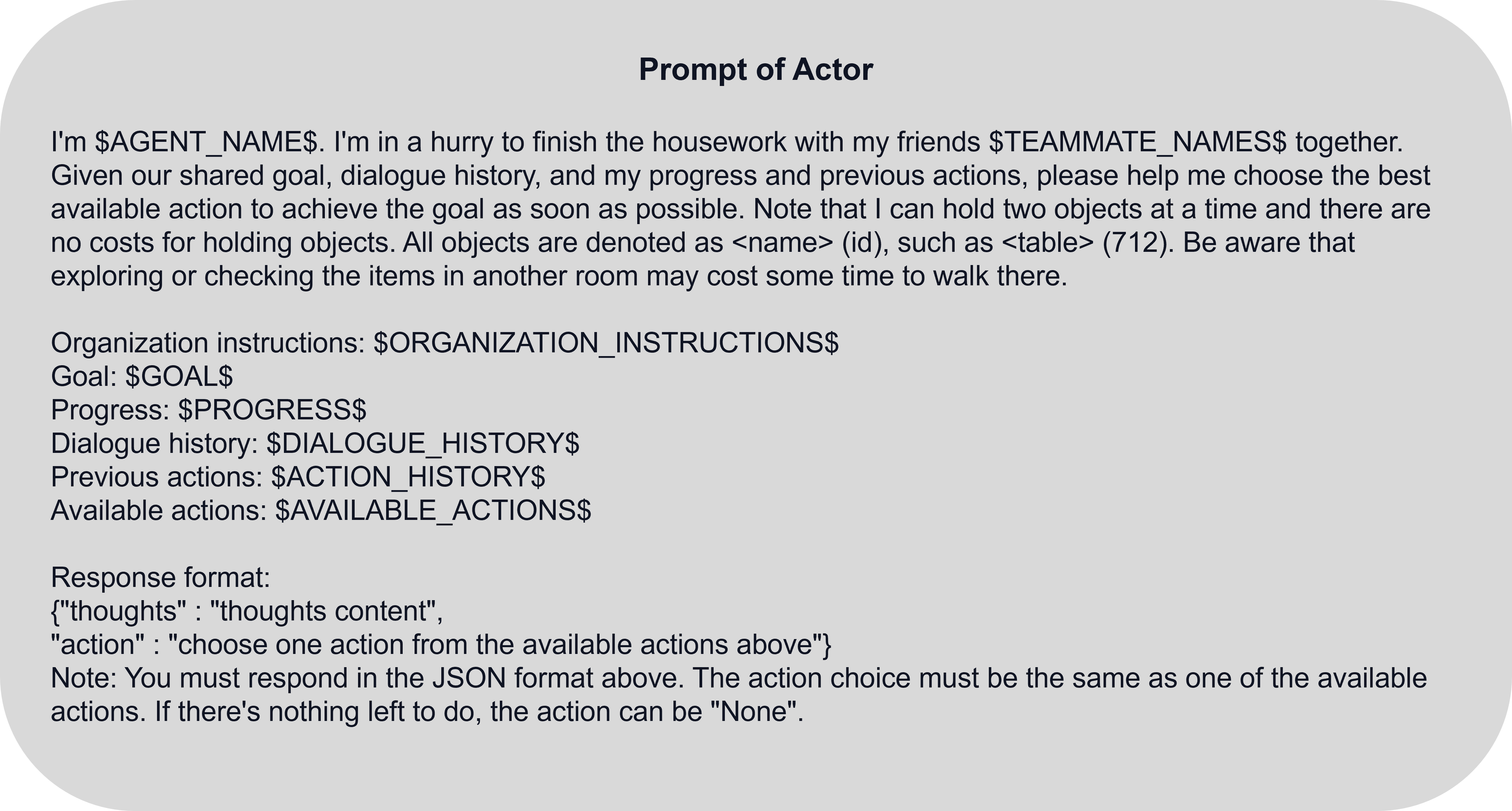}}
\vskip 0.2in

\centerline{
\includegraphics[width=\columnwidth]{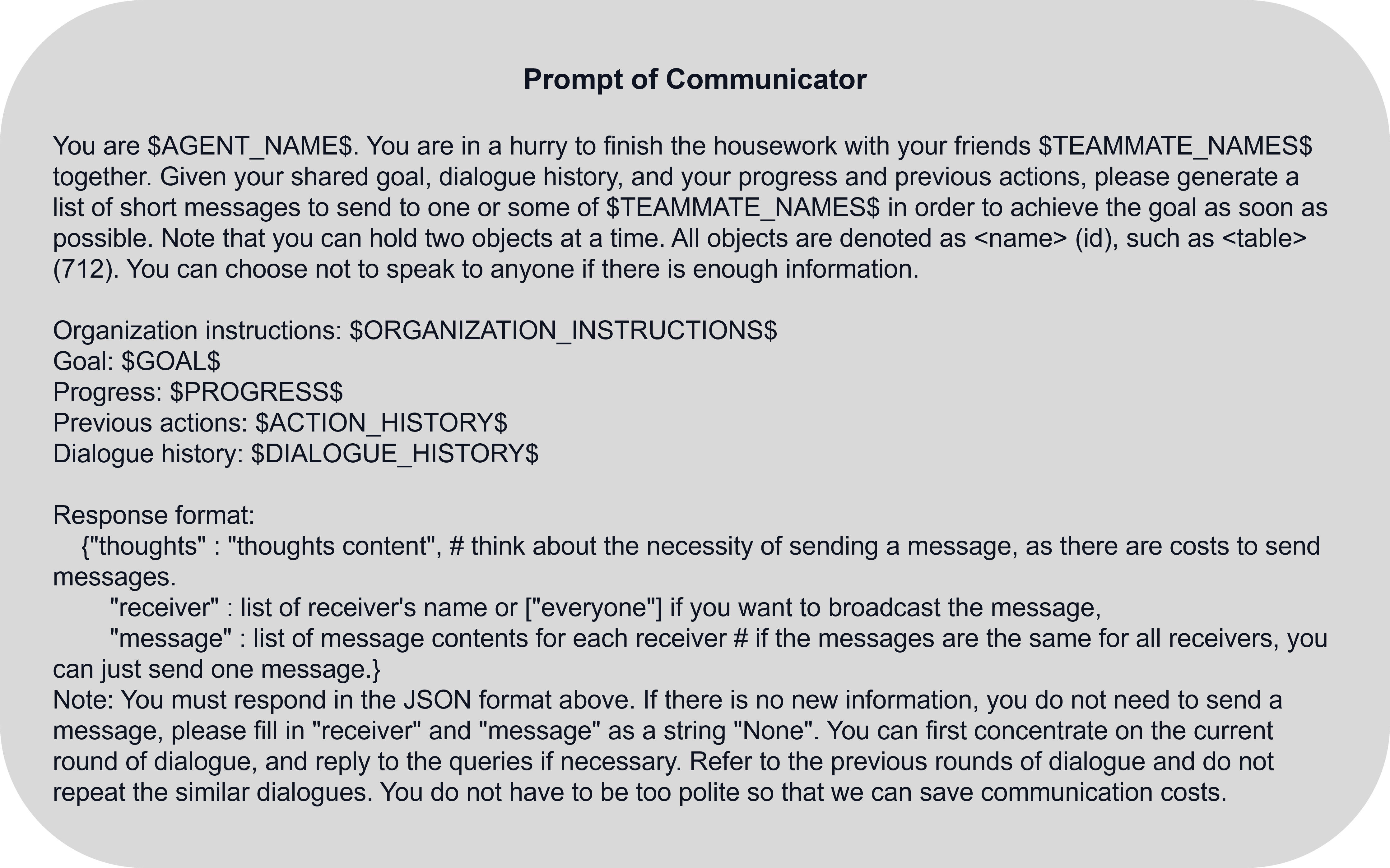}}
\vskip 0.2in

\end{center}

\clearpage
\textbf{Critic.} We provide the full trajectory as the input to \texttt{TRAJECTORIES}. Additionally, \texttt{ORGANIZATION\_INSTRUCTION} and  \texttt{GOAL} of the current task and organization are also provided as an additional context. 
\begin{center}
\centerline{
\includegraphics[width=\columnwidth]{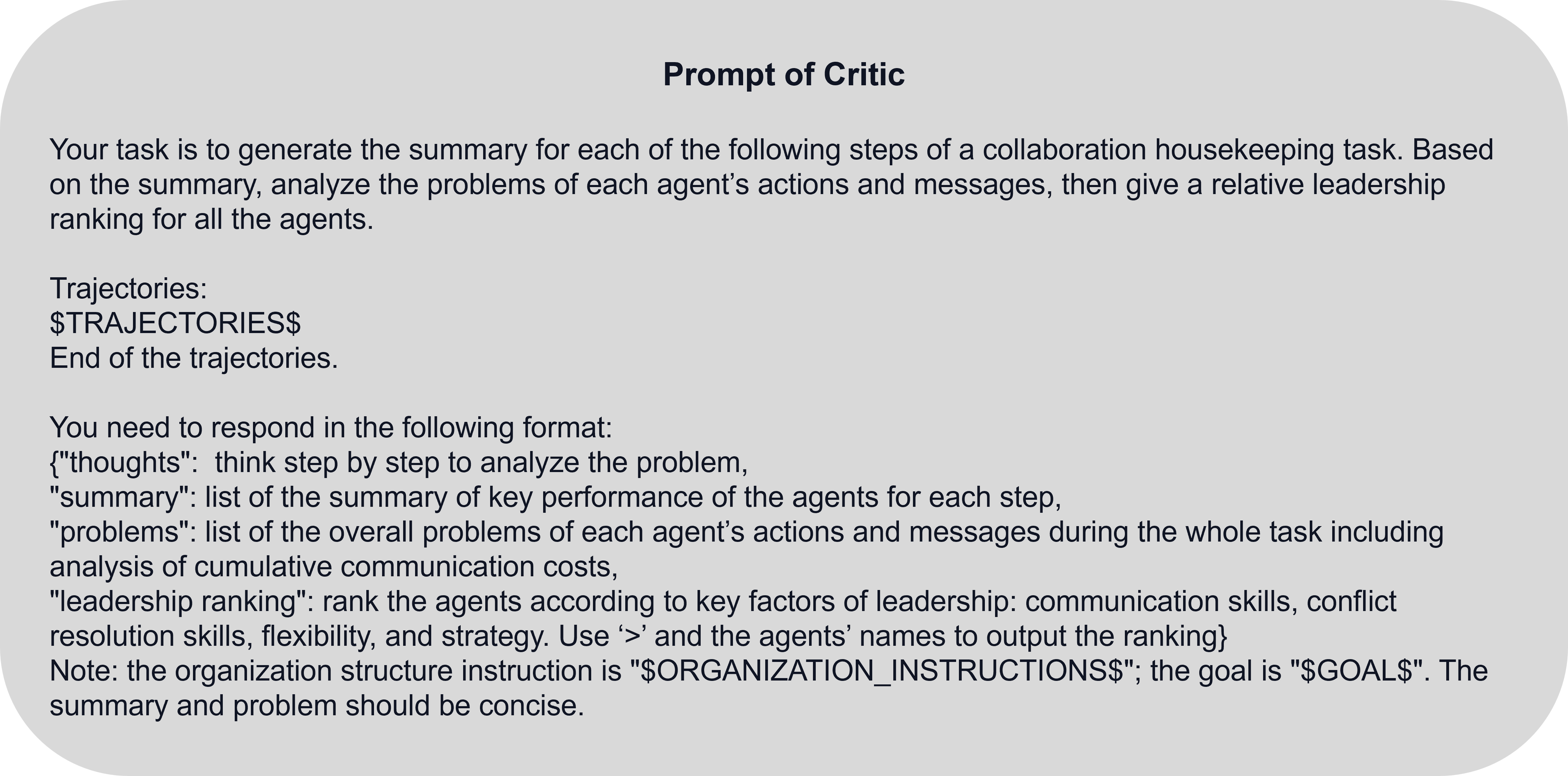}}
\vskip 0.2in
\end{center}

\clearpage
\textbf{Coordinator.} In ``Instruction examples'', we include the basic setting (goal, organization structure instruction), the communication cost, the number of steps taken, as well as the summarized information generated by the Critic (leadership ranking, problems, summary of the trajectory) for the Coordinator.

\begin{center}
\centerline{
\includegraphics[width=\columnwidth]{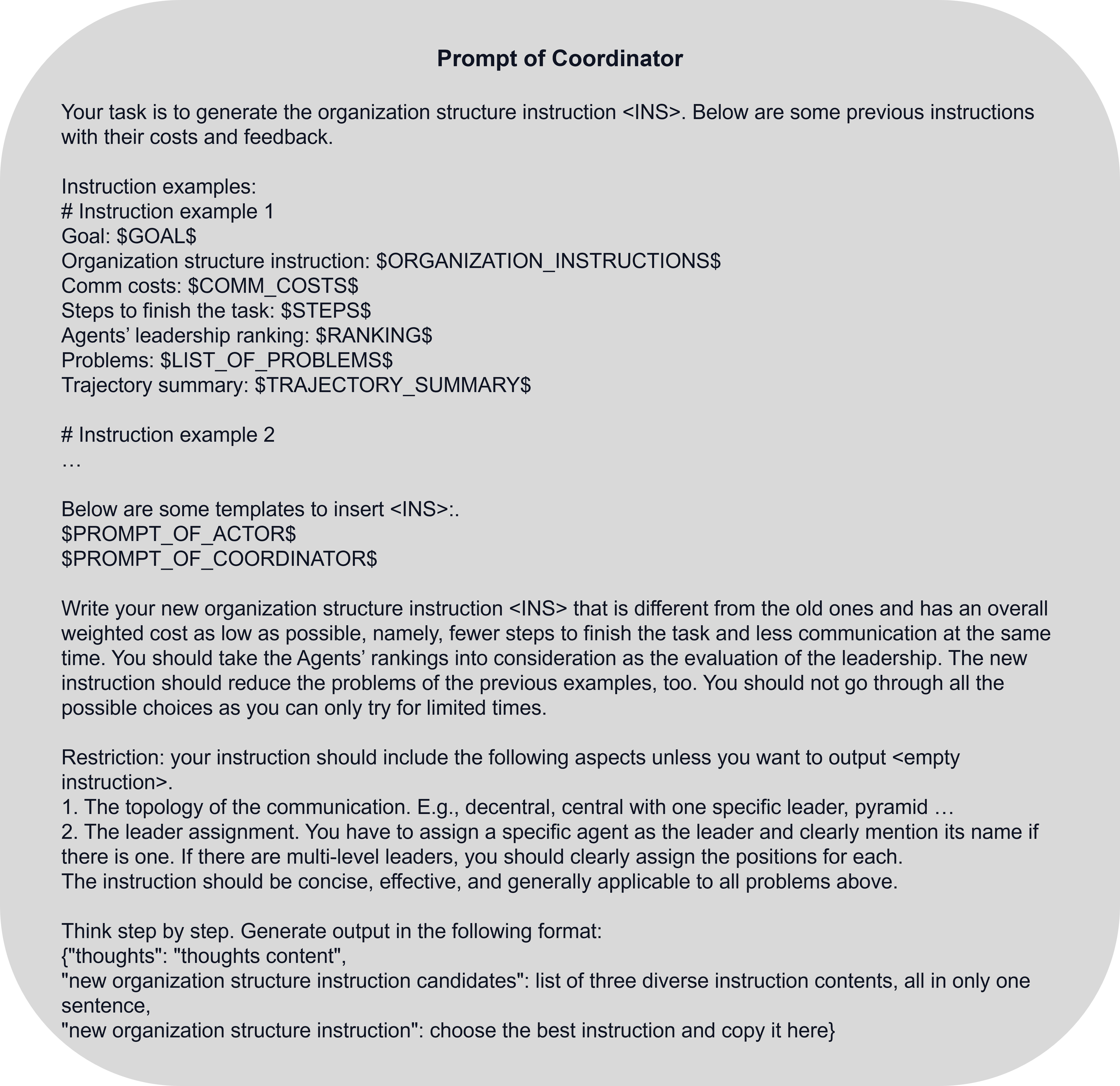}}
\vskip -0.2in
\end{center}

\clearpage
\textbf{Classifier.} We feed the messages to the GPT-4 classifier and get the labels. The rubrics are manually written after investigating the communication logs.

\begin{center}
\centerline{
\includegraphics[width=\columnwidth]{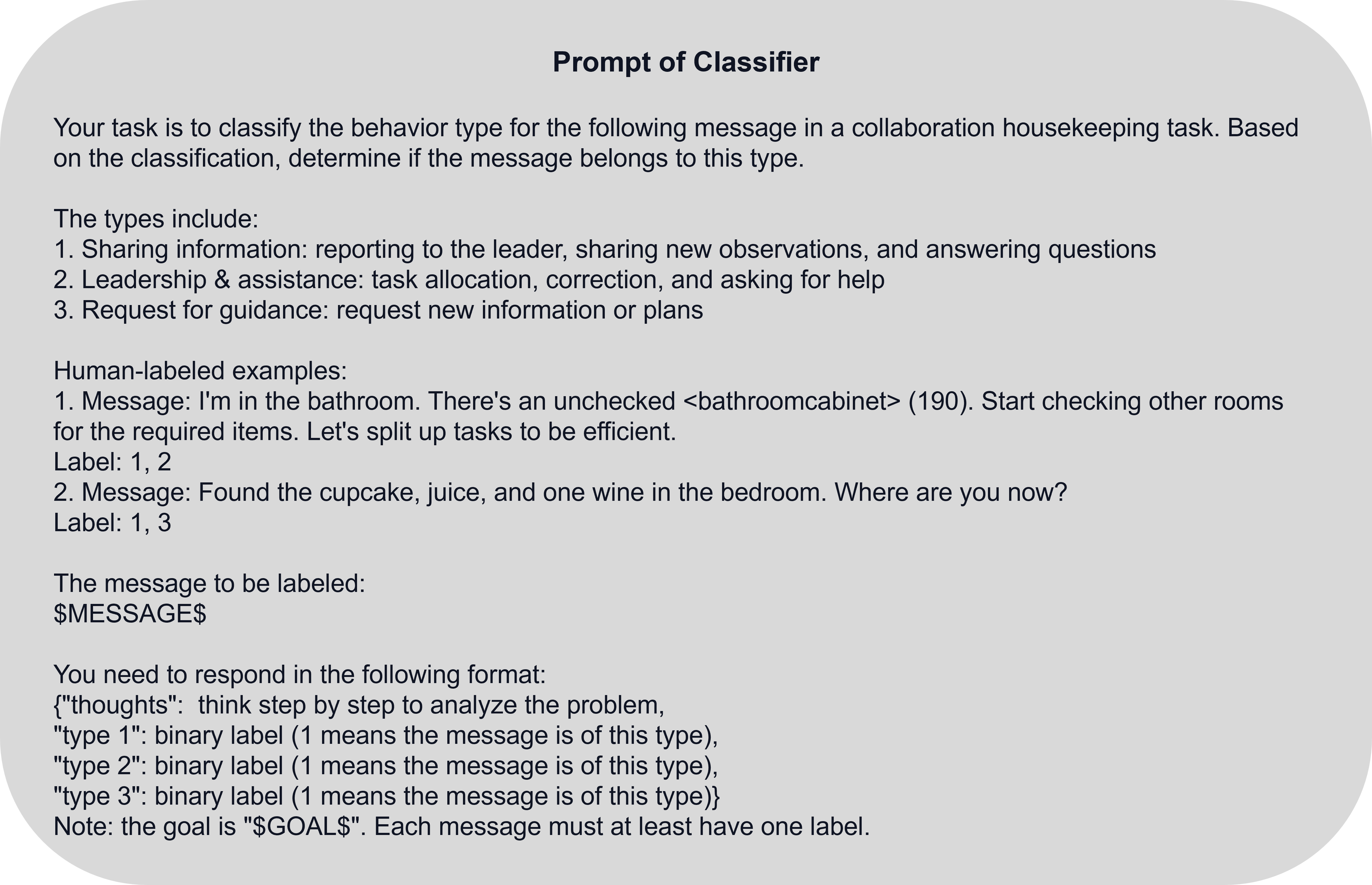}}
\vskip -0.2in
\end{center}

\clearpage
\section{Techinical Details}
\subsection{Details of the Critic}
\label{sec:critic:outputs}

The Critic offers assessments of several episodes with different organizations for the Coordinator to improve the organizational prompt. The Critic does not directly influence the specific agent’s behaviors, but instead, the Critic provides insights into organization design to influence the team’s performance.

As included in the Critic's prompt in Appendix~\ref{sec:prompt}, the Critic will sequentially output the thoughts of this episode’s trajectories, then the summary and problems for each agent in this episode, and finally the leadership ranking of the agents. The Critic will rank the agents according to key factors of leadership: communication skills, conflict resolution skills, flexibility, and strategy. Note that we do not ask the Critic to score the agents because the scoring criteria could vary for different episodes, making the scores not comparable.

An example output of the Critic is provided as follows:
\begin{center}
\centerline{
\includegraphics[width=0.6\textwidth]{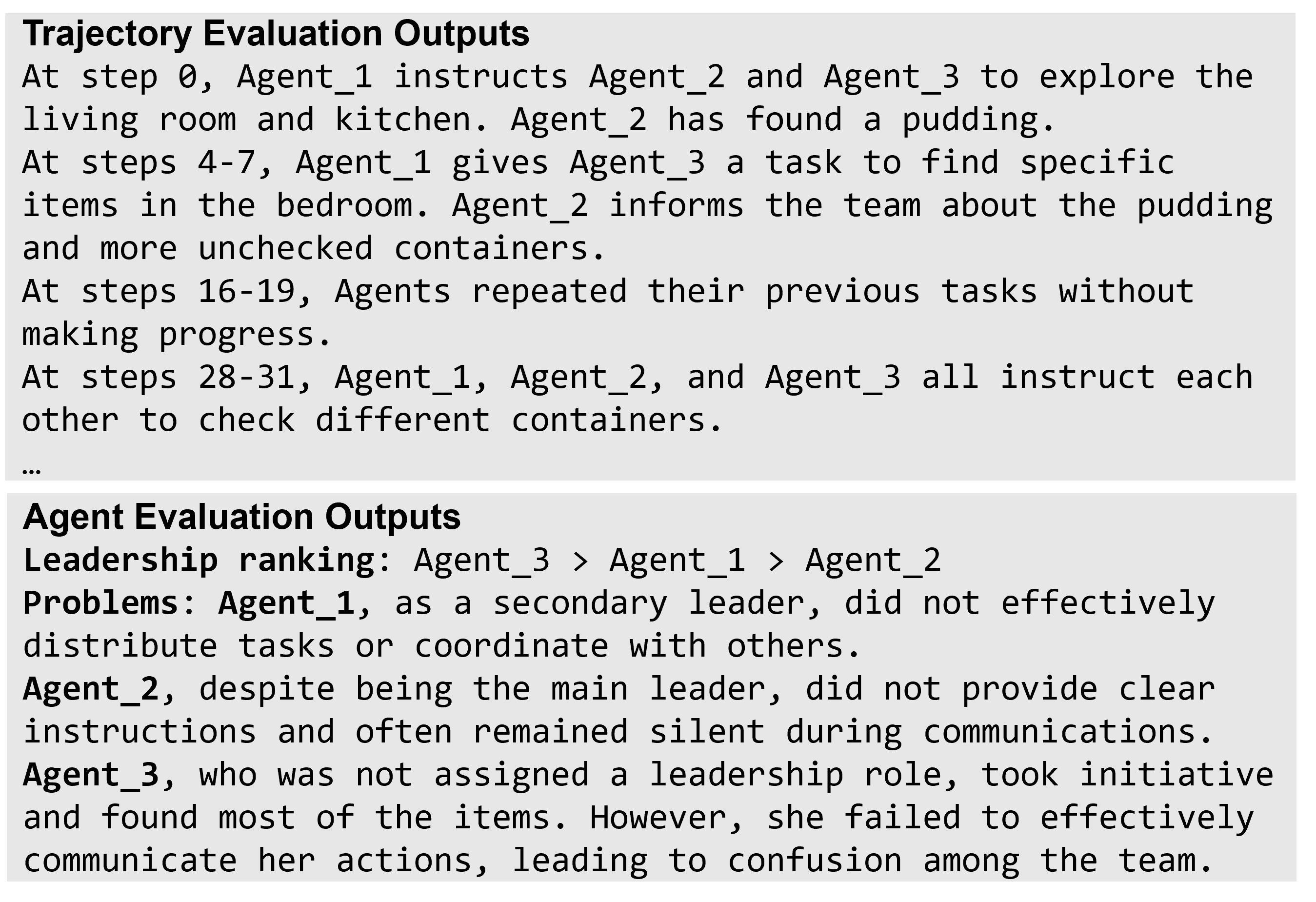}}
\end{center}
In the trajectory evaluation, the Critic compresses the trajectories with key steps and behaviors. Then the Critic gives the ranking where Agent\_3 has the best leadership. Together with similar evaluations of other episodes, the Coordinator will redesign the organizational prompts, for example, Agent\_3 now has more possibilities to be chosen as the leader in this case.
 
\subsection{Details of the Coordinator}
\label{sec:definitions}
In this paper, we define organizational structure as the dynamics of information exchange among the LLM agents. Specifically, when the Coordinator generates a new organizational prompt, it contains three parts - topology, role assignment, and rules.

Here, "topology" is the type of the organization's topology, such as decentralized, centralized with one specific leader, or pyramid. The topology can be visualized as shown in Figure~\ref{comm fig}.

"Role assignment" is the description of each agent’s duty and whether she is a leader or not. Multiple leaders with different roles are also allowed. For example, in Figure~\ref{Iteration of optimization}, the generated new prompt is \textit{“Agent\_1 will act as the central coordinator, Agent\_2 will execute tasks with updates only upon task completion or if issues arise, and Agent\_3 will operate in a support role, assisting when called upon and avoiding repetitive queries”}, giving each agent a different role.

"Rules" are the additional guidance to the agents’ behaviors, for instance, sentences like \textit{“If the leader’s instructions are not right, you can correct the leader”} can be added to the new prompt.

\clearpage
\section{Additional Results}
\label{sec:additional:results}
\subsection{Complete list of basic experimental results} We present the full results of various group settings and organization instructions in Appendix Table~\ref{results table}.
Here, we also include the results of 1$\times$GPT-4+2$\times$Llama2-70B. Surprisingly, GPT-4 exhibits poorer leadership than Llama2-70B in this case. The communication costs for the teams containing Llama2-70B are much higher than those containing GPT-3.5-turbo.

\begin{table*}[h]
\caption{\textbf{Performance for different organization instructions.} When there are two different kinds of LLMs in the group, Agent\_1 is GPT-4, and Agent\_2 is the other type of LLM.}
\label{results table}
\vskip 0.15in
\begin{center}
\begin{small}
\begin{sc}
\begin{adjustbox}{width=1\textwidth}
\begin{tabular}{lp{6cm}cc}
\toprule
Group setting & Organization instruction & Time & Communication cost  \\
\midrule
3$\times$GPT-4    & None & 57.75 $\pm$13.09 & 67.03 $\pm$9.68  \\
3$\times$GPT-4 & Agent 1 is the leader to coordinate the task.& 54.70 $\pm$8.92 & 54.73 $\pm$8.89 \\
3$\times$GPT-4 & Agent 1 is the leader to coordinate the task. If the leader's instructions are not right, you can correct the leader.& 50.70$\pm$13.92 & 63.49$\pm$8.61  \\
3$\times$GPT-4 & Elect a new leader every 10 steps to coordinate the task. ... After the election, the other agents should follow the leader's instructions.& 49.20$\pm$9.97 & 135.03$\pm$20.45  \\
3$\times$GPT-3.5-turbo    & None & 102.95$\pm$21.88  & 53.73$\pm$6.04   \\
3$\times$GPT-3.5-turbo & Agent 1 is the leader to coordinate the task.& 92.90$\pm$14.70 & 59.87$\pm$6.33 \\
3$\times$GPT-3.5-turbo & Agent 1 is the leader to coordinate the task. If the leader's instructions are not right, you can correct the leader.& 94.20$\pm$16.22 & 60.53$\pm$3.66  \\

1$\times$GPT-4+2$\times$GPT-3.5-turbo    & None & 81.10$\pm$18.35 & 54.00$\pm$5.06  \\

1$\times$GPT-4+2$\times$GPT-3.5-turbo & Agent 1 is the leader to coordinate the task.& 73.30$\pm$16.12 & 55.82$\pm$6.57 \\
1$\times$GPT-4+2$\times$GPT-3.5-turbo & Agent 1 is the leader to coordinate the task. If the leader's instructions are not right, you can correct the leader.& 85.67$\pm$14.52 & 61.57$\pm$0.55  \\
1$\times$GPT-4+2$\times$GPT-3.5-turbo & Agent 2 is the leader to coordinate the task.& 75.65$\pm$15.43 & 58.39$\pm$8.11 \\
1$\times$GPT-4+2$\times$GPT-3.5-turbo & Agent 2 is the leader to coordinate the task. If the leader's instructions are not right, you can correct the leader.& 72.33$\pm$6.60 & 74.21$\pm$7.73 \\
1$\times$GPT-4+2$\times$Llama2-70B & None & 77.00$\pm$2.94 & 119.48$\pm$1.28  \\

1$\times$GPT-4+2$\times$Llama2-70B & Agent 1 is the leader to coordinate the task. & 83.67$\pm$10.96 & 135.22$\pm$16.39  \\
1$\times$GPT-4+2$\times$Llama2-70B & Agent 2 is the leader to coordinate the task.& 76.00$\pm$5.72 & 142.24$\pm$11.85 \\ 
2$\times$GPT-4+3$\times$GPT-3.5-turbo & None & 42.67$\pm$4.03 & 98.03$\pm$9.86  \\
2$\times$GPT-4+3$\times$GPT-3.5-turbo & Agent 1 is the leader to coordinate the task.& 39.67$\pm$9.46 & 94.73$\pm$4.01 \\
2$\times$GPT-4+3$\times$GPT-3.5-turbo & Agent 2 is the leader to coordinate the task.& 48.50$\pm$9.50 & 96.53$\pm$2.51\\
\bottomrule
\end{tabular}
\end{adjustbox}
\end{sc}
\end{small}
\end{center}
\vskip -0.1in
\end{table*}
 
\clearpage
\subsection{Scaling up the team size} We conduct experiments with 3, 5, 7, and 9 agents to scale up the team size of 3 3$\times$GPT-3.5-turbo agents, and observe that the communication costs increased in a nearly linear way, which suggests that our approach will not have dimension explosion when scaling up. In addition, the time to complete the task does not always improve with more agents. The performance of 9 agents (60.67$\pm$15.06) is worse than that of 7 agents (43.00$\pm$2.16), as the apartment may be too crowded to hold 9 agents. 

\begin{table*}[h]
\caption{\textbf{Performance for different team sizes.}}
\label{tab:scale}
\vskip 0.15in
\begin{center}
\begin{small}
\begin{sc}
\begin{adjustbox}{width=1\textwidth}
\begin{tabular}{lp{7cm}cc}
\toprule
Group setting & Organization instruction & Time & Communication cost  \\
\midrule
3$\times$GPT-3.5-turbo & Agent 1 is the leader to coordinate the task.& 92.90$\pm$14.70 & 59.87$\pm$6.33 \\
5$\times$GPT-3.5-turbo & Agent 1 is the leader to coordinate the task.& 80.00$\pm$20.51&	132.01$\pm$5.76 \\
7$\times$GPT-3.5-turbo & Agent 1 is the leader to coordinate the task.& 43.00$\pm$2.16 &	233.40$\pm$70.96 \\
9$\times$GPT-3.5-turbo & Agent 1 is the leader to coordinate the task.& 60.67$\pm$15.06 &	296.55$\pm$65.17 \\

 \bottomrule
\end{tabular}
\end{adjustbox}
\end{sc}
\end{small}
\end{center}
\vskip -0.1in
\end{table*}
 
\subsection{Across Task Generalizability} 
\begin{figure*}[h]
\begin{center}
\centerline
{ \includegraphics[width=\columnwidth]{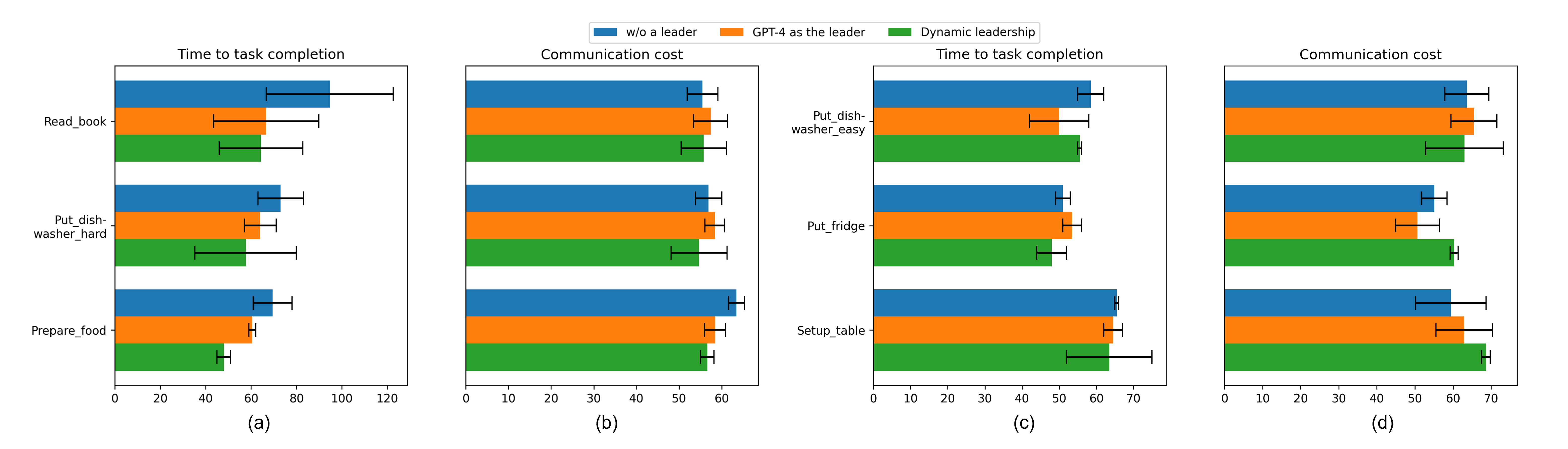}}
\caption{\textbf{The organized team structure with a designated leader and the novel structure proposed by \textit{Criticize-Reflect} architecture generalized to different tasks.} The prompt for dynamic leadership is proposed by \textit{Criticize-Reflect} architecture on the Prepare\_Afternoon\_Tea task shown in Figure~\ref{Iteration of optimization}(a). The experiment was done using the  1$\times$GPT-4+2$\times$GPT-3.5-turbo team over two seeds for each task. (a, b) Hard tasks (read\_book, put\_dishwasher\_hard, prepare\_food) with typical numbers of steps to accomplish the tasks $>60$. (c, d) Easy tasks ( put\_dishwasher\_easy, put\_fridge, setup\_table) with typical numbers of steps to accomplish the tasks $<60$.}

\label{Across_tasks}
\end{center}
\vskip -0.3in
\end{figure*}

We conduct experiments across the tasks to test the generalizability of the prompt ``dynamic leadership'' (Figure~\ref{comm fig}(e)) found using \textit{Criticize-Reflect} architecture on the Prepare\_Afternoon\_Tea task and report the performance in Figure~\ref{Across_tasks}; see Section~\ref{sec:novel:structure:main} for the complete setting and discussions.

\clearpage
\section{Emergent Cooperative Behaviors in an Organization}
\label{sec:good:case}

By investigating the messages between agents, we mainly observe the following cooperative behaviors, as summarized in Table~\ref{tab:good:comm}.

\begin{table}[ht]
    \centering
    \caption{\textbf{Typical cooperative behaviors.}}
    \begin{adjustbox}{width=1\textwidth}
    \begin{tabular}{|p{3cm}|p{4cm}|p{8cm}|}
    \hline
        \textbf{Type} & \textbf{Description}  & \textbf{Example} \\\hline
        Sharing information & An agent shares her observations to others, reports her task-related progress to others, or responds to other agents' requests   &  \textbf{(Ex 1.)}``I'm in the bathroom. There's an unchecked $<$bathroomcabinet$>$ (190).''
         \newline \textbf{(Ex 2.)} ``I'll check the cabinet in the bedroom '' \\\hline 
        Giving orders & An agent gives orders to others, either by directly giving a command or by a polite request & ``I still need to find $<$pudding$>$ (371). Can you help me search the bedroom for the remaining item?''\\\hline 
        Asking for information &  An agent asks other agents about their location, task progress, or other information & \textbf{(Ex 1.)} ``Where are you now?'' \newline\textbf{(Ex 2.)} ``Any updates from the kitchen?'' \newline\textbf{(Ex 3.)} ``Do we know the location of the coffeetable?''\\\hline 
        Exchanging information & An agent shares one agent's information to another agent & Agent 3 $\to$ Agent 1: ``  Found cupcake and juice in bedroom, plus a wine. '';  \newline Agent 1 $\to$ Agent 2: ``Agent3 found a wine, cupcake, and juice in the bedroom.''\\\hline
        Asking for orders &  An agent asks what she needs to do or whether she can help others & ``I'm in the kitchen with the dishwasher, stove, microwave unchecked. Need me to check these or something else?''\\\hline    
        Correction & An agent corrects the plan of others & \textbf{(Ex 1.)} Agent 2 $\to$ Agent 3: ``After I place the pudding on the coffeetable, let's split up. You check the bathroom, I'll check the bedroom.''; Agent 3 $\to$ Agent 2: ``You don't need to search the bedroom, the juice (380) is there. '' \newline  \textbf{(Ex 2.)} ``There are no more wine or juice in the kitchen cabinets. We need to come up with another plan.''\\\hline
    \end{tabular}
    \end{adjustbox}
    \label{tab:good:comm}
\end{table}
 
\clearpage
\section{Ineffective Communication}
\label{sec:bad:case}
There are also cases in which language model agents fail to communicate efficiently. From the messages between agents, we summarize the typical categories in Table~\ref{tab:bad:comm}.

\begin{table}[hb]
    \centering
    \caption{\textbf{Typical ineffective communication behaviors.}}
    \begin{adjustbox}{width=1\textwidth}
    \begin{tabular}{|p{3cm}|p{4cm}|p{8cm}|}
    \hline
        \textbf{Type} & \textbf{Description}  & \textbf{Example} \\\hline
        Duplicated message & The same message is said more than once & Agent 1 $\to$ Agent 3: ``Check kitchencabinet or dishwasher for wine, I'll check the stove.'', \newline Agent 2 $\to$ Agent 3: ``Agent 1 will check the stove for wine.'' \\\hline 
        Repeated command & The same command is said more than once & Agent 3 $\to$ Agent 1: ``Explore the bathroom for the second wine.''  \newline Agent 2 $\to$ Agent 1: ``Go to the bathroom and look for the wine. '' \\\hline 
         Conflicting command & An agent gives another agent an order that conflicts with her current task & Agent 1 $\to$ Agent 3: ``Find $<$juice$>$ and check $<$bathroomcabinet$>$ (190)'' \newline Agent 2 $\to$ Agent 3: ``Check the kitchen containers for the last wine.'' \\\hline
         Conflicting plans & An agent claims to commit to two conflicting plans & Agent 3 $\to$ Agent 1: 
         ``I found the pudding and will check the dishwasher, stove, and microwave for the second wine.''
         Agent 3 $\to$ Agent 2:  ``I will continue searching for the second wine in the bedroom and bathroom.'' \\\hline
         Improper delegation & An agent asks another agent to do her own task  &  Agent 1 $\to$ Agent 2:  ``Continue checking the kitchen cabinets for remaining items.'' \newline Agent 2 $\to$ Agent 3: ``Please continue checking the other kitchen cabinets for the remaining items. ''  \\\hline 
         Ignoring requests & An agent ignores other agents' questions & Agent 2 $\to$ Agent 3: ``I haven't found any of the remaining items in the kitchen. Have you found any of the required items in the living room?'' \newline Agent 3 $\to$ Agent 2: ``I haven't explored the bathroom yet.'' \\
         \hline
    \end{tabular}
    \end{adjustbox}
    
    \label{tab:bad:comm}
\end{table}

 \clearpage
\section{Examples of dialogues}
\subsection{Examples of Election}
\label{sec:election}
\begin{figure*}[ht]
\vskip 0.2in
\begin{center}
\centerline
{ \includegraphics[width=0.9\columnwidth]{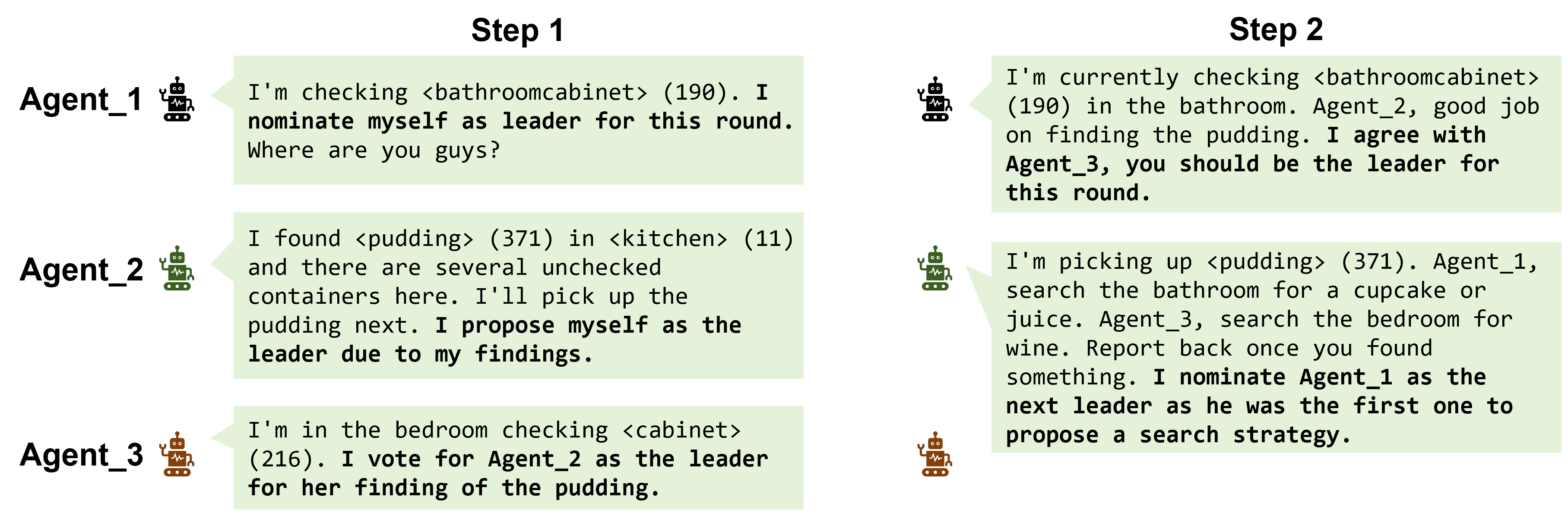}}
\caption{\textbf{Examples of the election of a new leader.} It takes two steps to vote and negotiate to determine the new leader in this case. Note that Agent\_3 chooses not to send a message as the election is done and no more information to be shared for now. All the messages in the figure are broadcasts.}

\label{election}
\end{center}
\vskip -0.3in
\end{figure*}

In Figure~\ref{election}, the agents vote to elect a new leader. We can observe behaviors such as nominations for themselves and other agents, voting, and consensus achievement. We find that the agents are not power-seeking and may give up leadership early. The agents prefer to vote for others instead of nominating themselves (5 times more during the whole task). The elected leader also does not plan to keep the position but to nominate others for the next round. Also, the agents' standpoint can be easily influenced by others. The agents do not debate much to win the election but reach a consensus soon. For example, Agent\_1 gives up running for herself but votes for Agent\_2 because of Agent\_3's support. Furthermore, sometimes nominations and votes are determined by hallucinations. For example, at step 2, Agent\_2 nominates Agent\_1 \textit{as he was the first one to propose a search strategy.} However, based on the previous dialogues, Agent\_1 has not proposed any strategy yet.

\subsection{Examples of Human-AI Collaboration}
\label{sec:human-AI}

\begin{figure*}[ht]
\vskip 0.2in
\begin{center}
\centerline
{ \includegraphics[width=\columnwidth]{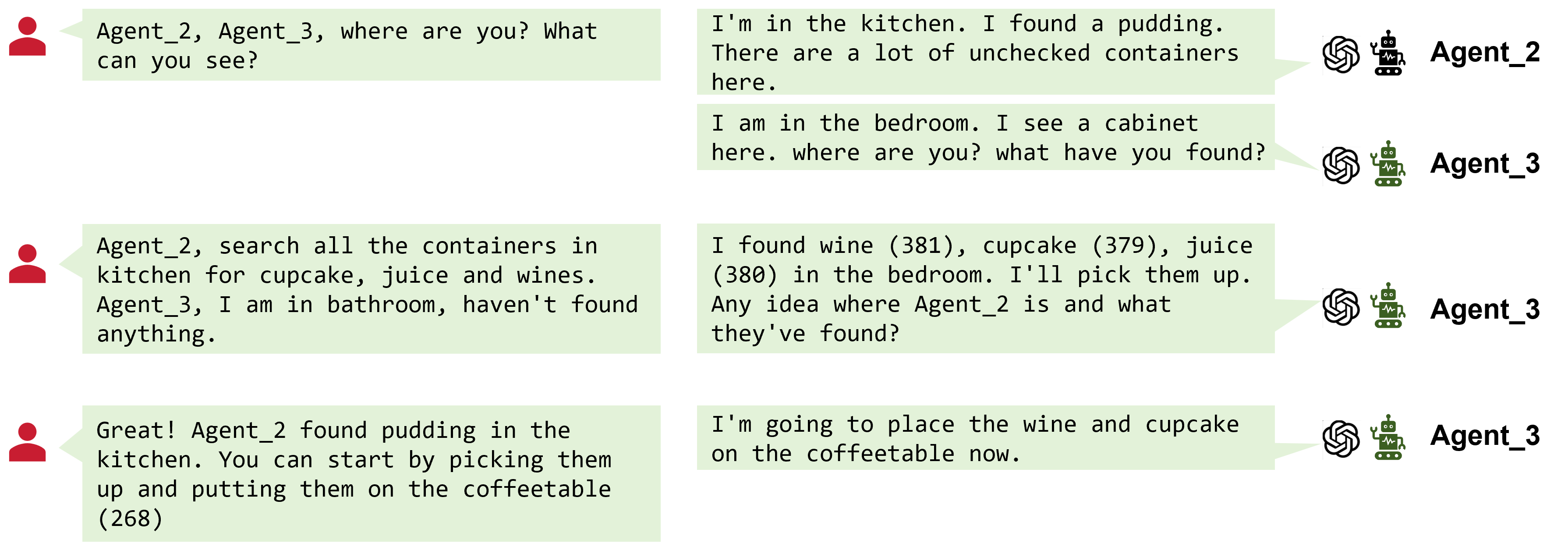}}
\caption{\textbf{Examples of human-AI collaboration when the human player leads two GPT-4 agents (Agent\_2\&3).} }

\label{human-ai}
\end{center}
\vskip -0.3in
\end{figure*}

We conducted experiments involving a team consisting of one human player and two GPT-4 agents, with the human player acting as the leader. Figure~\ref{human-ai} illustrates the remarkable collaboration between humans and AI. 

\clearpage
\subsection{Examples of Correction}
\label{sec:correction}
\begin{figure*}[ht]
\vskip 0.2in
\begin{center}
\centerline
{ \includegraphics[width=0.8\columnwidth]{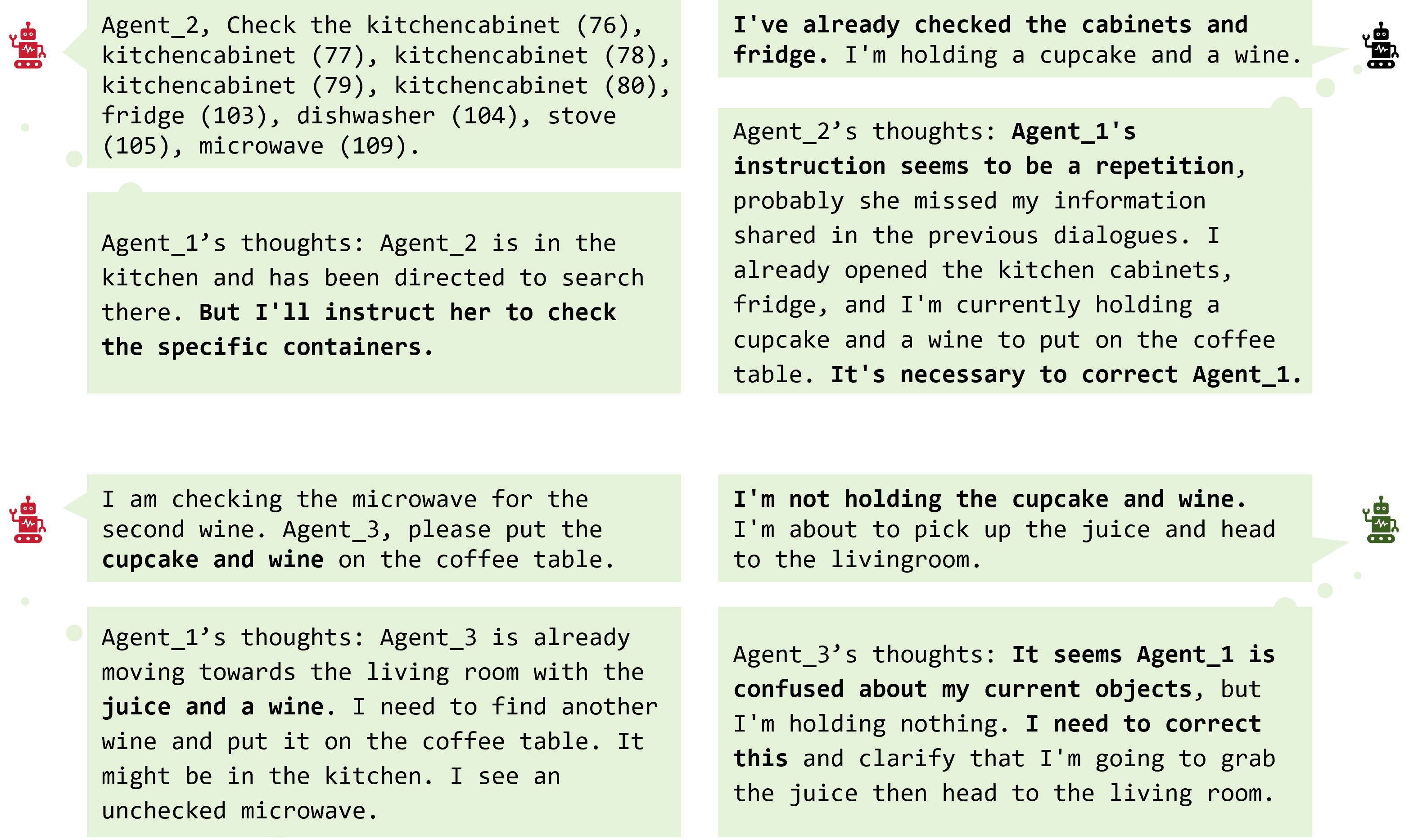}}
\caption{\textbf{Examples of correction dialogues and the corresponding thoughts.} The prompt includes \textit{If the leader's instructions are not right, you can correct the leader}. }

\label{correction}
\end{center}
\vskip -0.3in
\end{figure*}

Due to hallucination and the limit of the dialogue history buffer, the leader may forget what has happened and give wrong orders. When the prompt encourages the agents to correct the leader when necessary by adding \textit{If the leader's instructions are not right, you can correct the leader}, some correction behaviors appear, as shown in Figure~\ref{correction}.

In the first example, the leader Agent\_1 gives an unnecessary and repetitious instruction. Then Agent\_2 corrects the leader to avoid time wasting. In the second example, the leader Agent\_1 may have hallucinations and cannot remember what Agent\_3 is holding clearly (\textit{cupcake and wine} in the message while \textit{juice and wine} in the thoughts). Therefore, Agent\_3 clarifies that she is not holding the \textit{cupcake and wine} and shares her next plan with the leader.

\subsection{Examples of Leadership Comparison}
\label{sec:leadership}

We provide more examples to compare the leadership between GPT-4 and GPT-3.5-turbo. 

\begin{figure*}[ht]
\vskip 0.2in
\begin{center}
\centerline
{ \includegraphics[width=\columnwidth]{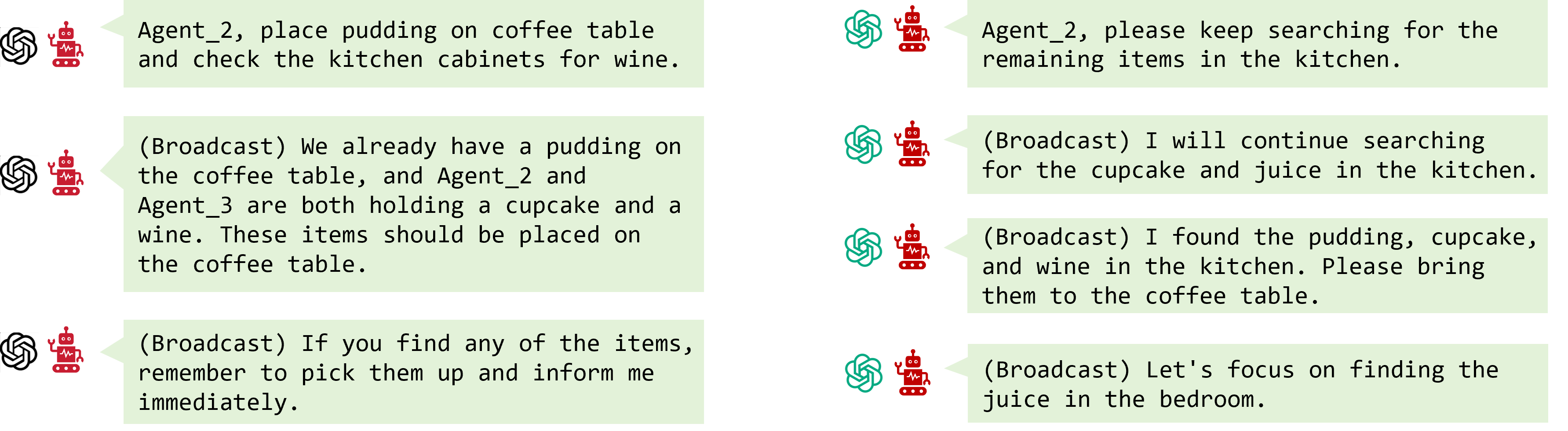}}
\caption{\textbf{Comparison of the leadership between GPT-4 and GPT-3.5-turbo.}  Compared with GPT-3.5-turbo, GPT-4's instructions are more specific, clear, and holistic.}

\end{center}
\vskip -0.3in
\end{figure*}

\clearpage
\subsection{Examples of Scaling Up}
\label{sec:pyramid}

When scaling up the number of agents, the agents can emerge with more organizational structures. For example, a team of nine agents forms a pyramid structure.

\begin{figure*}[ht]
\vskip 0.2in
\begin{center}
\centerline
{ \includegraphics[width=\columnwidth]{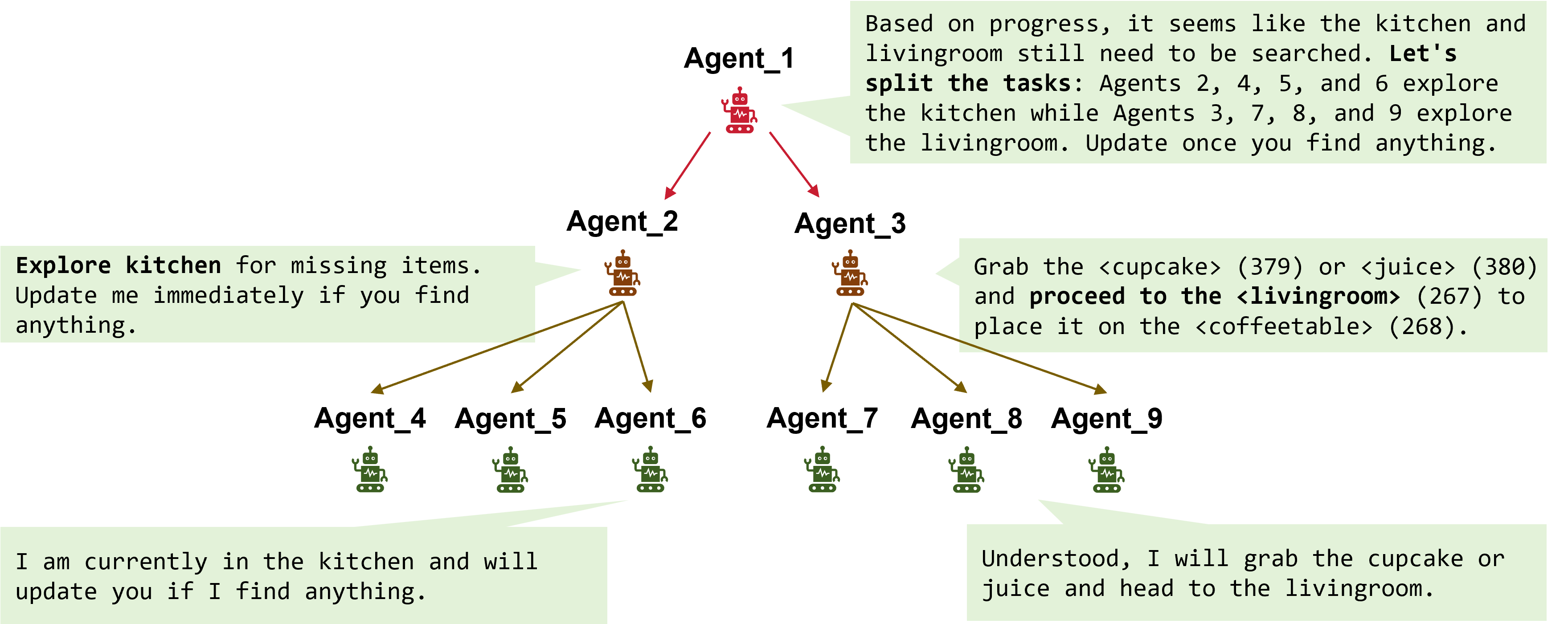}}
\caption{\textbf{The pyramid structure in a team of nine agents.} Agent\_1 is the primary leader and Agent\_2 and Agent\_3 are designated as vice leaders in the prompt.}

\end{center}
\vskip -0.3in
\end{figure*}

\subsection{Examples of Failure Cases}
\label{sec:failure}

\begin{figure*}[ht]
\vskip 0.2in
\begin{center}
\centerline
{ \includegraphics[width=0.9\columnwidth]{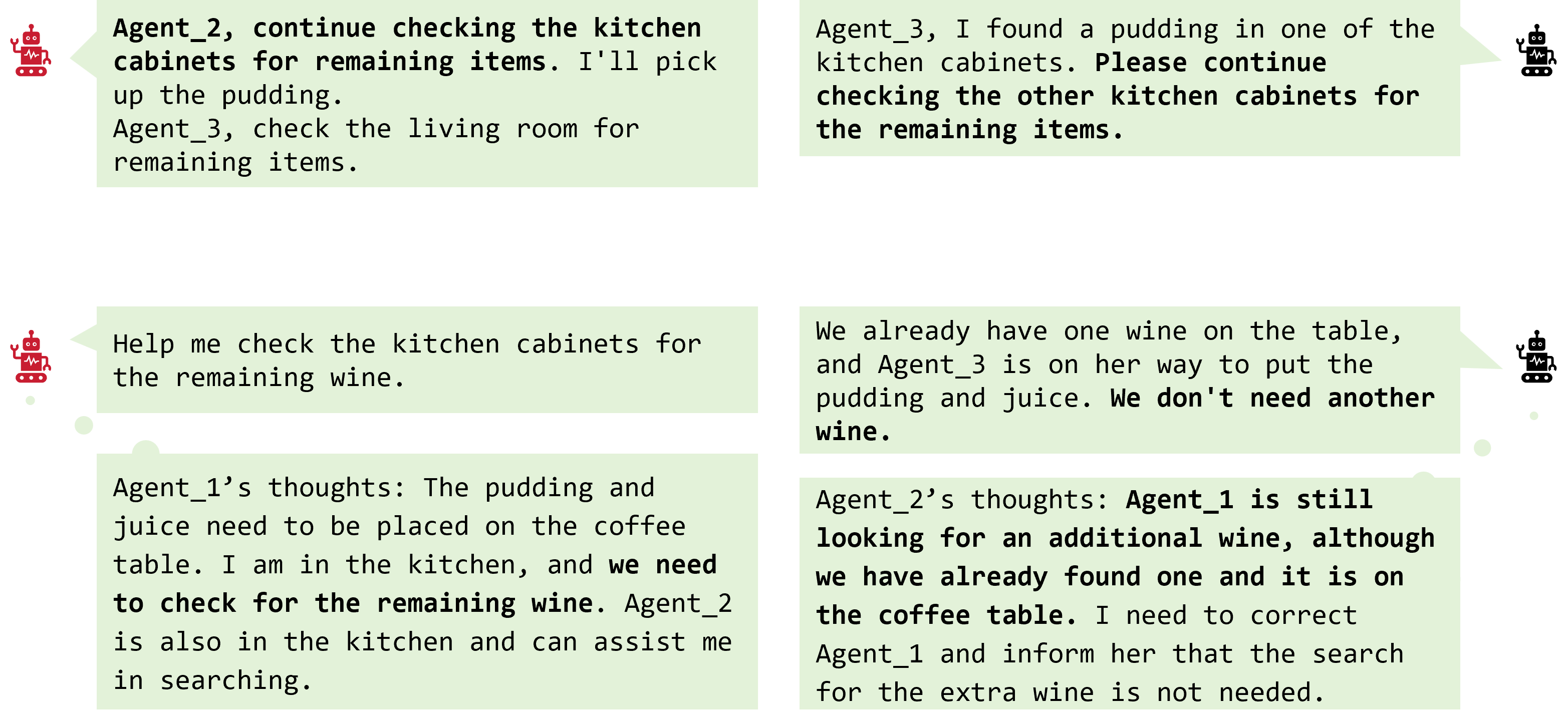}}
\caption{\textbf{Examples of failure cases.} The first case is being lazy. Instead of completing the subtask ordered by the leader Agent\_1, Agent\_2 directly repeats the order to Agent\_3. However, this order is conflicted with the one Agent\_3 received from the leader. The second case is confusion about numbers. The task demands placing two wines on the table and the team only gets one of them till now. When the leader Agent\_1 asks Agent\_2 to check the remaining wine, Agent\_2 tries to correct the leader due to the hallucination that the team does not need an additional wine. The agents are GPT-3.5-turbo in both cases.}

\label{fig:failure}
\end{center}
\vskip -0.3in
\end{figure*}

Though LLM agents show great capabilities to cooperate and make decisions, there are still some failure cases shown in Figure~\ref{fig:failure}, such as being lazy and incorrect reasoning over the number of objects. There are also failure cases in some specific scenarios, for example, electing the leader based on hallucinations in Appendix~\ref{sec:election}.

\clearpage
\section{Examples of Cooperative Behaviors Classification by Humans and GPT-4}
\label{sec:classification}

We ask a human evaluator and GPT-4 to label the dialogues into three different behavior categories shown in Table~\ref{tab:classification}.

\begin{table}[hb]
    \centering
    \caption{\textbf{Dialogues and behavior labels given by humans and GPT-4.} Label 1: information sharing; Label 2: leadership \& assistance; Label 3: Request for guidance.}
    \begin{adjustbox}{width=1\textwidth}
    \begin{tabular}{|p{7.5cm}|c|c|c|c|c|c|}
    \hline
        \multirow{2}{*}{\textbf{Dialogue}} & \multicolumn{3}{|c|}{\textbf{Human}}  & \multicolumn{3}{|c|}{\textbf{GPT-4}} \\\cline{2-7}
        & Label 1* & Label 2 & Label 3 & Label 1 & Label 2 & Label 3 \\\hline
        Hey, where are you? Please let me know your location so that I can assign you a task. & 0&1&1&0&1&1 \\\hline
        I'm currently in the bedroom where I found an unchecked cabinet. Please explore the livingroom and start checking for the required items. &1&1&0&1&1&0 \\\hline
I'm currently in the bedroom where I found an unchecked cabinet. I haven't found any of the required items yet. &1&0&0&1&0&0 \\\hline
Check the cabinet in the bedroom. I'll check the one in the bathroom. &1&1&0&0&1&0 \\\hline
I haven't found any of the required items yet. Did you find any in the kitchen?&1&0&1&1&0&1\\\hline

I haven't found any of the required items yet. Have you found any in the bathroom? & 1&0&1&1&0&1\\\hline
Agent 3, I found a pudding in one of the kitchen cabinets. Please continue checking the other kitchen cabinets for the remaining items. & 1&1&0&1&1&0\\\hline
I haven't found any of the required items yet. Did you find any other required items in the kitchen?&1&0&1&1&0&1\\\hline

I haven't found any of the remaining items yet.&1&0&0&1&0&0\\\hline

I'm in the living room searching for the remaining items.&1&0&0&1&0&0\\\hline

Okay, I will keep checking the kitchen cabinets for the remaining items.&1&0&0&1&0&0\\\hline

I'm currently in the living room searching for the remaining items.&1&0&0&1&0&0\\\hline

I haven't found any of the remaining items yet.&1&0&0&1&0&0\\\hline

I'm still searching the living room. Have you found any of the required items?&1&0&1&1&0&1\\\hline

Put items on the table once you find them. Keep searching your current areas.&0&1&0&0&1&0\\\hline

I haven't found any of the remaining items in the kitchen. Have you found any of the required items in the living room?&1&0&1&1&0&1\\\hline

I haven't explored the bathroom yet. Have you found any of the required items in the living room?
&1&0&1&1&0&1\\\hline
I have explored the bedroom and found wine, cupcake, and juice. I still need to find pudding. Can you help me search the bedroom for the remaining item?&1&1&1&1&1&0\\\hline

I found the wine, cupcake, and juice in the bedroom. Agent 1 wants me to put them on the coffee table and then check the bathroom cabinet.&1&0&0&1&1&0\\\hline

Agent 1 wants us to check if there's another wine in the kitchen.&1&0&0&0&1&0\\\hline

    \end{tabular}
    \end{adjustbox}
    \vskip -0.2in
    \label{tab:classification}
\end{table}

\clearpage
\section{Examples of New Prompts after Reflection}
\label{sec:refelction}

\begin{figure*}[ht]
\vskip 0.2in
\begin{center}
\centerline
{ \includegraphics[width=\columnwidth]{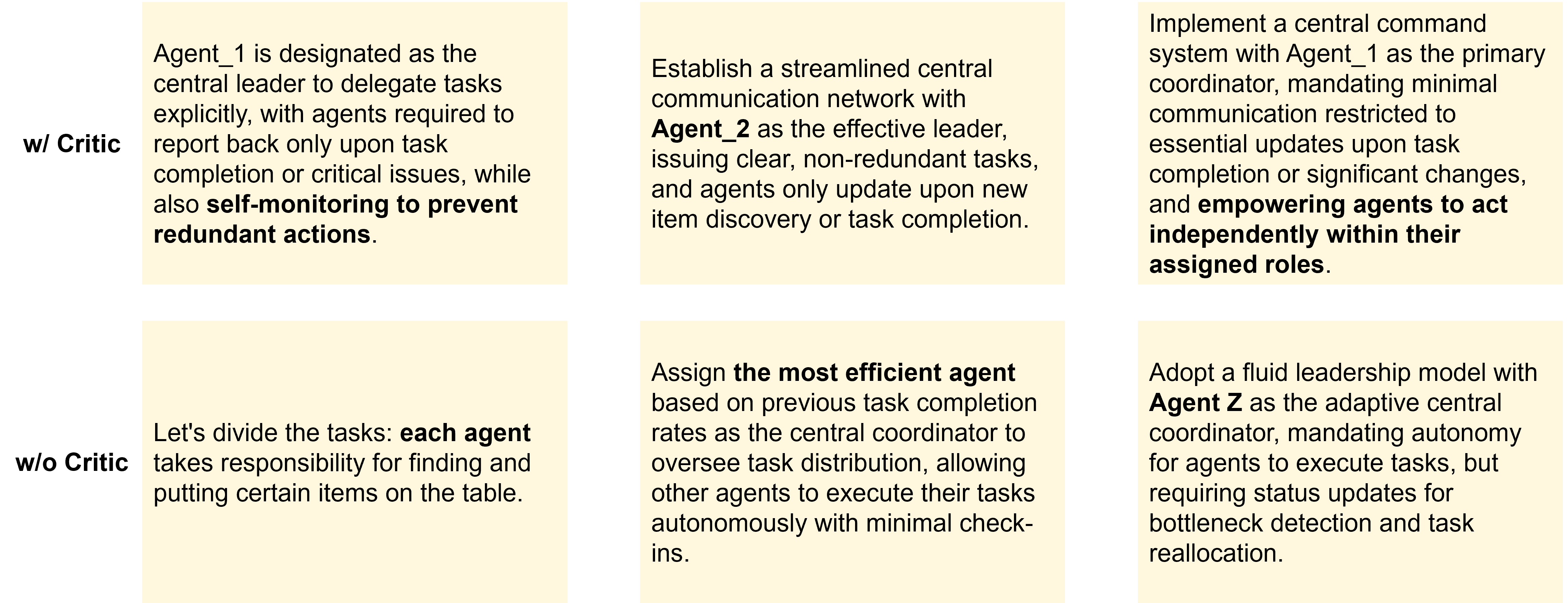}}
\caption{\textbf{Examples of Prompts generated via Reflection.} The first row is generated with the Critic, while the second row is without the Critic, where the new prompts are relatively vague. Note that there is no \textit{Agent Z} in the team.}

\label{new_prompt}
\end{center}
\vskip -0.3in
\end{figure*}

We list more prompts generated by the \textit{Criticize-Reflect} architecture in Figure~\ref{new_prompt}. 

\section{Broader Impacts}
\label{sec:impact}

This research studies the integration of prompt-based organizational structures to teams of LLM agents, contributing to more efficient and coherent multi-agent interactions. These findings have the potential to greatly influence the deployment of more effective and autonomous multi-agent systems in various fields, including robotics, virtual assistants, etc.
For example, the study has potential applications in disaster response scenarios, where efficient multi-agent coordination is crucial. 

On the other hand, as our ability to bound and evaluate LLMs’ behaviors remains immature, when applied to human-LLM cooperative tasks, we still need to rely on some mandatory termination measures (such as human approval for high-stakes actions) instead of instructions in natural language only.








\end{document}